\def\simpletoucharxiv{1}
\documentclass{article}
\usepackage[T1]{fontenc}
\usepackage{iclr2027_conference,times}
\usepackage{fontspec}

\fontencoding{T1}\selectfont
\usepackage{amsmath,amssymb,graphicx,booktabs,array,xcolor,colortbl}
\usepackage{wrapfig}
\usepackage{placeins}
\usepackage{capt-of}
\usepackage{hyperref,url}
\colorlet{TaclentCite}{cyan}
\colorlet{TaclentLink}{magenta}
\hypersetup{
  colorlinks=true,
  citecolor=TaclentCite,
  urlcolor=TaclentLink,
  filecolor=TaclentLink,
  linkcolor=TaclentLink,
  pdfborder={0 0 0}
}

\newfontfamily\simpletouchfont[Scale=0.88]{Optimistic.ttf}
\DeclareRobustCommand{\simpletouch}{{\simpletouchfont SimpleTouch}}

\newcommand{\pihalf}{\ensuremath{\pi_{0.5}}}

\hypersetup{pdftitle={SimpleTouch: Can Vision-Language-Action Models Master Contact-Rich Manipulation Without Tactile Policy Pretraining?}}

\title{SimpleTouch: Can Vision-Language-Action Models Master Contact-Rich Manipulation Without Tactile Policy Pretraining?}
\ifdefined\simpletoucharxiv
  \iclrfinalcopy
\hypersetup{
  pdfauthor={Chen Yang, Linzhe Shi, Changjie Wu, Hang Zhang, Ronghan Chen, Lingjun Zhang, Xu Hu, Mu Xu, Jiansheng Fan, Chen Wang},
  pdfsubject={Tactile-enhanced vision-language-action models for contact-rich manipulation},
  pdfkeywords={SimpleTouch, tactile manipulation, vision-language-action models}
}
\author{%
  Chen Yang\textsuperscript{1,2,*}\quad
  Linzhe Shi\textsuperscript{2,*}\quad
  Changjie Wu\textsuperscript{2}\quad
  Hang Zhang\textsuperscript{2}\quad
  Ronghan Chen\textsuperscript{2}\\
  \bfseries Lingjun Zhang\textsuperscript{1,2}\quad
  Xu Hu\textsuperscript{2,3}\quad
  Mu Xu\textsuperscript{2}\quad
  Jiansheng Fan\textsuperscript{1}\quad
  Chen Wang\textsuperscript{1,\ensuremath{\dagger}}\\[0.4em]
  {\normalfont\textsuperscript{1}Tsinghua University\quad
    \textsuperscript{2}Amap, Alibaba Group\quad
    \textsuperscript{3}Hong Kong Polytechnic University}%
}
\date{}
\makeatletter
\gdef\@thanks{%
  \footnotetext[1]{Equal contribution.\quad
    \textsuperscript{\ensuremath{\dagger}}Corresponding author: Chen Wang
    (\href{mailto:chwang@tsinghua.edu.cn}{\nolinkurl{chwang@tsinghua.edu.cn}}).}%
}
\renewcommand{\@maketitle}{%
  \vbox{\hsize\textwidth
    {\LARGE\sc\@title\par}
    \def\And{\end{tabular}\hfil\linebreak[0]\hfil
      \begin{tabular}[t]{l}\bf\rule{\z@}{14pt}\ignorespaces}%
    \def\AND{\end{tabular}\hfil\linebreak[4]\hfil
      \begin{tabular}[t]{l}\bf\rule{\z@}{14pt}\ignorespaces}%
    \begin{tabular}[t]{l}\bf\rule{\z@}{14pt}\@author\end{tabular}%
    \vskip 5.4pt
  }%
}
\makeatother
\renewenvironment{abstract}{%
  \par\vspace{5pt}%
  \begingroup
  \setlength{\parskip}{0pt}%
  \setlength{\topsep}{0pt}%
  \setlength{\partopsep}{0pt}%
  \centerline{\large\sc Abstract}%
  \nobreak\vspace{6pt}%
  \begin{quote}%
}{%
  \par\end{quote}%
  \endgroup
  \vspace{-4pt}%
}

\fi

\newcommand{\mainfigcaptionspace}{\setlength{\abovecaptionskip}{4pt}}

\begin{document}
\maketitle
\vspace{-0.8em}
\begin{center}
\begin{minipage}{\linewidth}
\centering
\makeatletter
\def\@captype{figure}
\makeatother
\ifdefined\simpletoucharxiv
\includegraphics[width=0.95\linewidth]{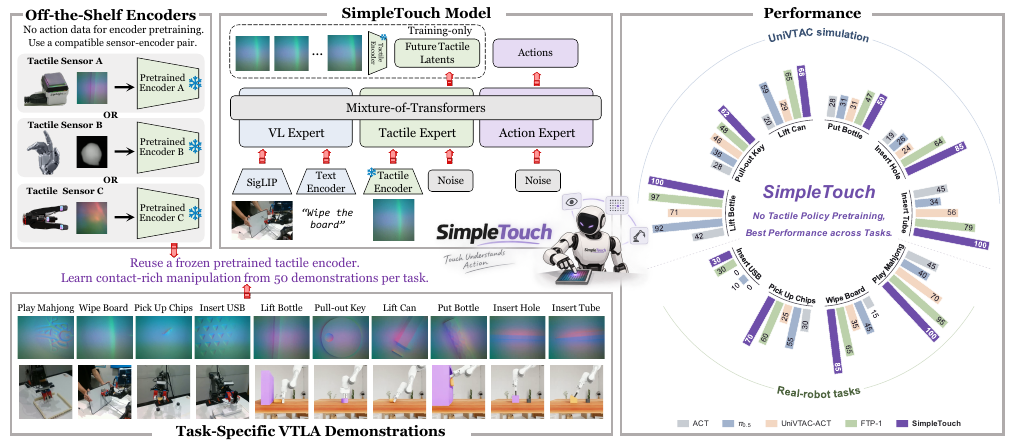}
\else
\includegraphics[width=\linewidth]{figures/simpletouch_teaser.pdf}
\fi
\mainfigcaptionspace
\caption{Overview of \simpletouch{}. \simpletouch{} combines an off-the-shelf pretrained tactile encoder with a tactile expert to learn from 50 task-specific demonstrations without tactile policy pretraining. The chart on the right reports success rates on six UniVTAC and four real-robot tasks.}
\label{fig:teaser}
\end{minipage}
\end{center}

\begin{abstract}
Tactile sensing provides essential contact information for robotic manipulation, yet incorporating it into pretrained vision-language-action (VLA) models remains challenging.
A common concern is that simply introducing touch during task-specific fine-tuning may fail to bridge the cross-modal gap, yielding limited gains or even reduced success.
Consequently, existing methods often rely on large-scale tactile policy pretraining or separate visuotactile alignment, adding data requirements and training stages.
We introduce \simpletouch{}, a simple VLA extension that augments \pihalf{} with a tactile expert, to test whether these additional stages are necessary.
Leveraging all tokens from a frozen pretrained tactile encoder, the expert learns from action supervision and multi-horizon prediction of future tactile latents.
This single-stage training uses only task demonstrations, without additional tactile policy pretraining or separate alignment.
With 50 demonstrations per task, \simpletouch{} achieves the highest success rate among evaluated methods on all six UniVTAC tasks.
Its average success rate reaches 77.5\%, compared with 45.2\% for FTP-$\pi_{0.5}$ and 66.7\% for FTP-1, corresponding to gains of 32.3 and 10.8 percentage points, respectively.
Across four real-world tasks, it averages 71.3\%, exceeding FTP-1 by 8.8 percentage points.
These results demonstrate that, given pretrained VLA and tactile representations, additional tactile policy pretraining is not a prerequisite for strong performance on these tasks, offering a simpler route to contact-rich manipulation.
\ifdefined\simpletoucharxiv
Project page: \url{https://simpletouch-robot.github.io/}
\fi

\end{abstract}

\section{Introduction}

Robot manipulation has made substantial progress through learned policies~\citep{zhao2023act,chi2023diffusionpolicy} and pretrained vision-language-action (VLA) models~\citep{kim2025openvla,black2024pi0}. Much of this progress has relied on visual perception as the primary interface for scene understanding. However, visual observations do not fully determine contact-rich interaction: two manipulation states can appear visually similar while requiring different actions because their contact geometry, deformation, or slip state differs~\citep{li2014gelsight,yuan2017gelsight}. Tactile sensing supplies local feedback about these otherwise hidden contact states, motivating policies that learn from touch and use it during manipulation~\citep{xue2025rdp,huang2025tactilevla,lou2026dreamtac}.

However, adding touch to a pretrained policy does not automatically make it useful. Tactile observations lie outside the pretraining modalities of many VLA backbones, and directly incorporating them during fine-tuning can yield limited gains or even reduce success~\citep{niu2026trex,li2026atvla,zhang2026tacvla}. The central challenge is therefore not merely obtaining tactile features, but presenting them to a pretrained VLA in a form that preserves contact information while allowing the policy to learn their task-specific use from limited demonstrations. Representative responses include large-scale tactile policy pretraining, as in FTP-1~\citep{yuan2026ftp}, and separate visuotactile representation alignment, as in VITaL~\citep{george2024vital}. These routes provide useful priors but introduce additional data requirements and training stages. Meanwhile, pretrained tactile encoders already offer reusable contact representations~\citep{zhao2025t3}. This distinction between acquiring tactile representations and learning a tactile policy motivates our question: \emph{given a pretrained VLA and tactile encoder, can a policy learn strong contact-rich manipulation from task-specific tactile demonstrations without additional tactile policy pretraining or a separate alignment stage?}

We introduce \simpletouch{}, a simple recipe that augments \pihalf{}~\citep{intelligence2025pi05} with one tactile expert, without additional tactile policy pretraining or a separate alignment stage. Its design follows two principles. First, it preserves the pretrained tactile representation: we freeze the tactile encoder and retain all current tokens, including global and spatial features, as a stable source of contact information. Second, it learns the task-specific use of that representation through a tactile expert. The tactile expert processes current tactile features separately from the visual-language backbone, and the action expert reads them through layerwise attention. During joint training, the expert learns from both action supervision and multi-horizon prediction of future tactile latents through latent flow matching~\citep{lipman2023flow}. This task-trained pathway gives action generation direct access to high-fidelity tactile features. During task-level training, we jointly optimize the pretrained \pihalf{} policy, including its visual-language backbone and action expert, together with the tactile expert, while keeping the tactile encoder frozen.

With 50 demonstrations per task, \simpletouch{} achieves the highest success rate among the evaluated methods on all six UniVTAC~\citep{chen2026univtac} tasks. Its mean success rate of 77.5\% exceeds those of FTP-$\pi_{0.5}$ and the pretrained FTP-1 policy by 32.3 and 10.8 percentage points, respectively. Across four real-world tasks, \simpletouch{} averages 71.3\%, surpassing FTP-1 by 8.8 points and showing improvements in real-world experiments as well. At deployment, the tactile expert processes the current observation once, and its layerwise cache is reused throughout action-flow integration; future tactile prediction is used only as training supervision. Together, these results show that, given strong pretrained VLA and tactile representations, a frozen tactile representation and task-trained tactile expert can achieve strong performance without additional tactile policy pretraining or explicit alignment.

\section{Related Work}
\label{sec:related}

\paragraph{Generalist Policy Learning.}
Generalist policy learning converts broad visual and language priors into continuous robot control across diverse tasks. The $\pi$ series couples pretrained vision-language backbones with flow-matching action experts~\citep{black2024pi0,intelligence2025pi05}. Other VLA systems investigate large-scale data, efficient fine-tuning, execution, generalization, and modular policy development~\citep{oneill2023openx,khazatsky2024droid,bjorck2025groot,pertsch2025fast,kim2025oft,cai2026xiaomi,chen2025internvla,community2026starvla}. Complementary world-action models incorporate predictive visual or feature dynamics into action learning, including Video Prediction Policy, LingBot-VA, Motus, DreamZero, LaWAM, and Fast-WAM~\citep{hu2024vpp,li2026lingbot,zhang2026lingbot2,bi2025motus,ye2026dreamzero,chen2026lawam,yuan2026fastwam}. These directions establish strong reusable architectures and priors for action generation, but their observation interfaces are primarily built around vision, language, and proprioception rather than local tactile contact feedback.

\paragraph{Tactile Policy Learning.}
Tactile policy learning can be organized by where contact-sensitive priors are acquired. One route first learns how tactile observations participate in control through large-scale tactile policy pretraining or tactile-grounded mid-training, as in FTP-1, the $N_0$-TWAM/$N_0$-VTLA family, and T-Rex~\citep{yuan2026ftp,neote2026n0twam,neote2026n0vtla,niu2026trex}. A second route aligns tactile and visual representations before policy learning, including VITaL, UniTouch, and VT-MUSE~\citep{george2024vital,yang2024unitouch,xu2026vtmuse}. Both front-load resources beyond task-level learning. Other methods explore task-level tactile reasoning, adaptive fusion, or reactive feedback~\citep{yu2023mimictouch,huang2024threedvitac,heng2025vitacformer,liu2025vitamin,huang2025tactilevla,bi2025vlatouch,li2026atvla,xue2025rdp}, while predictive approaches model future contact or multimodal dynamics~\citep{lou2026dreamtac,niu2026touchdreaming,zheng2026omnivta}. Across these settings, simply adding tactile input during fine-tuning does not consistently improve success~\citep{yuan2026ftp,niu2026trex,li2026atvla}. An open question is whether strong contact-rich control instead requires an additional tactile policy-learning or alignment stage once pretrained VLA and tactile representations are available.

\paragraph{Pretrained Tactile Encoders.}
Pretrained tactile encoders address representation acquisition rather than action generation. Early work such as T-Dex learns self-supervised tactile embeddings from robotic play for downstream dexterous control~\citep{guzey2023tdex}. Sparsh and Sparsh-X learn general-purpose representations from self-supervised vision-based and multisensory touch data, respectively~\citep{higuera2024sparsh,higuera2025sparshx}; the AnyTouch series models static and dynamic tactile signals across sensors and downstream tasks~\citep{feng2025anytouch,feng2026anytouch2}. T3 combines sensor-specific encoders with a shared Transformer trained across sensors and tasks~\citep{zhao2025t3}. Such encoders provide reusable contact-sensitive features across tasks and sensing configurations. This differs from tactile policy pretraining: encoders supply sensory representations, whereas policy pretraining also learns how touch influences action generation. Existing encoder work does not prescribe the downstream action interface or establish whether a separate tactile policy-training stage remains necessary. A detailed review of related work is provided in Appendix~\ref{app:extended_related_work}.

\begin{figure}[t]
    \centering
    \includegraphics[width=\linewidth]{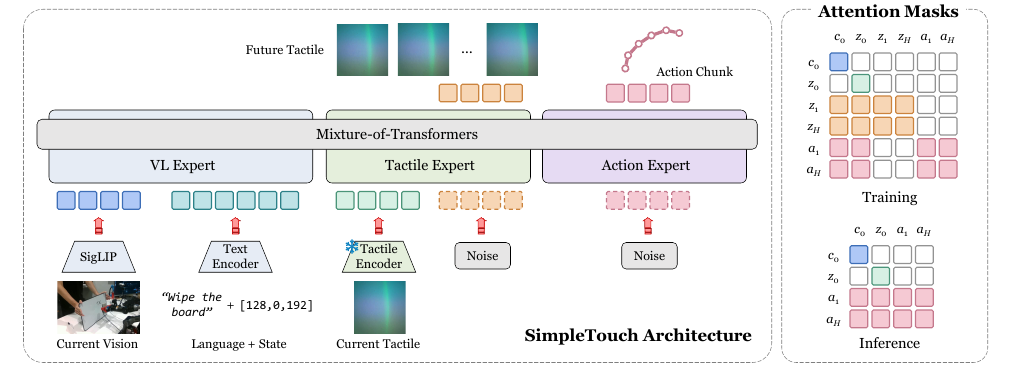}
    \mainfigcaptionspace
    \caption{\simpletouch{} architecture. A tactile expert uses complete current tactile features from a frozen encoder and learns multi-horizon future prediction. In the right-side training and inference masks, rows are queries, columns are keys/values, and colored cells mark permitted attention. The symbols $c_0$, $z_0$, $z_{1:H}$, and $a_{1:H}$ denote the current visual-language and state context, current tactile tokens, future tactile tokens, and action tokens, respectively, following the token order shown above.}
    \label{fig:method_overview}
    \vspace{-0.5em}
\end{figure}

\section{Method}
\label{sec:method}

\subsection{Problem Formulation}
\label{sec:problem}
We study task-level learning of contact-rich manipulation from a pretrained vision-language-action (VLA) policy, a pretrained tactile encoder, and task-specific tactile demonstrations. The demonstrations form trajectories $\mathcal D$ of synchronized visual observations, language instructions, proprioceptive states, tactile observations, and actions. At environment step $t$, let $O_t=(I_t,\ell,s_t,U_t)$ denote these current observations, where $I_t$, $\ell$, $s_t$, and $U_t$ are visual observations, a language instruction, proprioceptive state, and tactile observations, respectively. The policy predicts an action chunk $A_t=(a_t,\ldots,a_{t+H-1})$ over horizon $H$:
\begin{equation}
    A_t\sim\pi_\Theta(\,\cdot\mid I_t,\ell,s_t,U_t).
    \label{eq:policy}
\end{equation}
Our goal is to learn this policy using $\mathcal D$ alone, without an additional tactile policy-pretraining stage or a separate visuotactile alignment stage. The trainable parameters are $\Theta=\{\theta_{\mathrm{VL}},\theta_A,\theta_T\}$ for the visual-language backbone, action expert, and tactile expert, respectively; the tactile encoder parameters $\phi$ are held fixed during task-level training. Each trajectory also contains future tactile observations, which provide training-only supervision for the tactile expert.

Figure~\ref{fig:method_overview} summarizes the design. \simpletouch{} preserves the complete current representation of a pretrained tactile encoder, routes it through a tactile expert, and uses future tactile prediction as a training-time objective for learning the tactile-action pathway. The pretrained VLA policy continues to supply visual-language and action priors; the tactile expert provides the task-trained interface from touch to action. Implementation details of \simpletouch{} are provided in Appendix~\ref{app:implementation}.

\subsection{Preserving Pretrained Tactile Features}
\label{sec:pretrained_features}
The first design principle is to preserve, rather than re-learn, the pretrained tactile representation. We use pretrained T3~\citep{zhao2025t3} as a tactile encoder $E_\phi$ and keep its weights frozen. For the current tactile observation, it produces
\begin{equation}
    Z_t=E_\phi(U_t).
    \label{eq:tactile_encoding}
\end{equation}
We retain every global and spatial token in $Z_t$ without pooling or learned compression, thereby preserving the spatial layout encoded by T3. Thus, the downstream policy can access the encoder's full current representation of contact, while task-level training only learns how that representation should be used for control. Future tactile targets are treated differently: they are pooled only for predictive supervision, not for the current tactile context used by the policy.

\subsection{A Tactile Expert as a Tactile--Action Interface}
\label{sec:tactile_action_interface}
The \pihalf{} policy~\citep{intelligence2025pi05} comprises a visual-language backbone and an action expert. We augment it with a tactile expert of the same Transformer architecture as the action expert but with independent parameters. Rather than inserting tactile tokens into the visual-language backbone, its current-touch computation $T_{\theta_T}^{\mathrm{current}}$ processes $Z_t$ and produces the layerwise key-value cache
\begin{equation}
    \mathcal K_t=\{(K_{T,t}^{j},V_{T,t}^{j})\}_{j=1}^{L}
    =T_{\theta_T}^{\mathrm{current}}(Z_t),
    \label{eq:tactile_cache}
\end{equation}
where $L$ is the number of Transformer layers and $j$ indexes them. Let $C_t=B_{\theta_{\mathrm{VL}}}(I_t,\ell,s_t)$ be the visual-language and state context produced by the pretrained backbone. At each action-expert layer, action queries read visual-language context, current tactile context, and noisy action tokens:
\begin{equation}
    R_A^j=\operatorname{Attn}\!\left(Q_A^j,
    [K_C^j;K_{T,t}^j;K_A^j],
    [V_C^j;V_{T,t}^j;V_A^j]\right).
    \label{eq:expert_readout}
\end{equation}
Here $\operatorname{Attn}$ is multi-head attention, and $Q_A^j$ and $R_A^j$ are the action-block queries and outputs. For $X\in\{C,T,A\}$, $K_X^j,V_X^j$ are the layer-$j$ projections of visual-language context, current tactile features, and noisy action tokens, respectively; semicolons denote token concatenation. Expert-specific projections make the tactile keys and values compatible with the action expert. This layerwise interface exposes the full tactile token sequence throughout action generation while allowing tactile processing to use its own parameters. The tactile expert consequently learns a task-specific transformation from pretrained tactile features to action-relevant context.

\begin{figure}[t]
    \centering
    \includegraphics[width=0.85\linewidth]{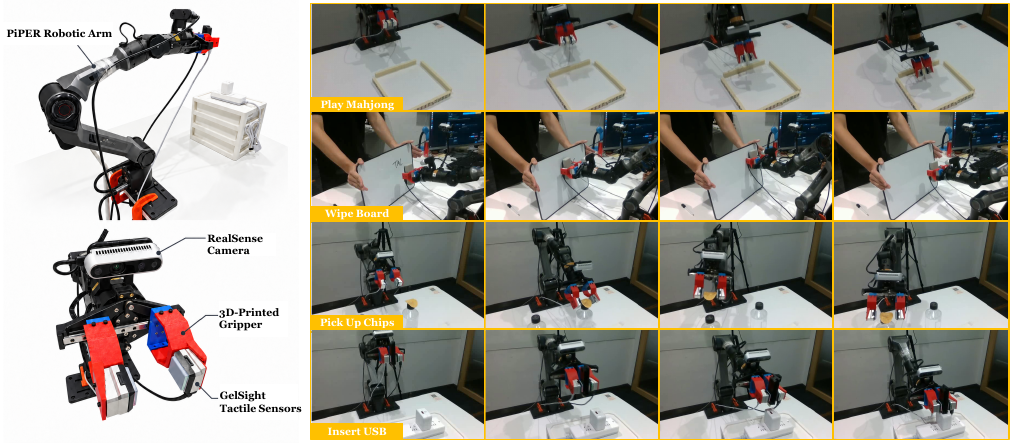}
    \mainfigcaptionspace
    \caption{Real-world setup and rollouts. The left panel shows the PiPER platform and sensing hardware; the right panel shows all four tasks---Play Mahjong, Wipe Board, Pick Up Chips, and Insert USB---with time progressing from left to right throughout each rollout.}
    \label{fig:real_rollouts}
    \vspace{-0.5em}
\end{figure}

\subsection{Learning Contact Dynamics from Future Touch}
\label{sec:predictive_supervision}
Action supervision alone trains the tactile expert only through the downstream behavior objective. We add multi-horizon future tactile prediction as dense supervision of contact evolution. For offsets $\Delta$ within the action horizon, future tactile observations are encoded by the same encoder and form the target sequence
\begin{equation}
    Y_t=\big[\mathcal S\big(\mathcal P(E_\phi(U_{t+\delta}))\big)\big]_{\delta\in\Delta}.
    \label{eq:future_target}
\end{equation}
Here $\mathcal P$ retains global tokens and spatially pools patch tokens, while $\mathcal S$ applies fixed feature scaling. This asymmetry preserves all tokens for current control while keeping the multi-horizon prediction target tractable.

Following the Fast formulation of Fast-WAM~\citep{yuan2026fastwam}, we treat future touch as an auxiliary prediction target rather than an intermediate condition for action generation:
\begin{equation}
    \underbrace{p_\Theta(A_t,Y_t\mid O_t)}_{\text{task-level training}}
    =
    \underbrace{\pi_\Theta(A_t\mid O_t)}_{\text{action generation}}
    \underbrace{p_\Theta(Y_t\mid O_t)}_{\text{future-touch prediction}}.
    \label{eq:training_factorization}
\end{equation}
Unlike future-conditioned formulations $p(Y_t\,|\,O_t)\,p(A_t\,|\,O_t,Y_t)$, the action factor depends only on current observations. We implement this separation with asymmetric attention: current tactile queries read current touch; future tactile queries read visual-language context, current touch, and noisy future tactile tokens; and action queries read current context but not future tactile tokens.

Both action chunks and future tactile targets are learned with conditional flow matching~\citep{lipman2023flow}. The joint objective is
\begin{equation}
    \mathcal L=\mathcal L_{\mathrm{act}}+\lambda_T\mathcal L_{\mathrm{tac}},
    \label{eq:loss}
\end{equation}
where both terms are flow-matching velocity losses and $\lambda_T$ weights tactile prediction. The current and future paths share tactile-expert parameters, so $\mathcal L_{\mathrm{tac}}$ trains the same expert that supplies current tactile context for action generation. We jointly optimize $\Theta=\{\theta_{\mathrm{VL}},\theta_A,\theta_T\}$.

At deployment, the tactile expert processes current touch once, and its layerwise cache $\mathcal K_t$ is reused throughout action-flow integration alongside the visual-language context. The auxiliary future-prediction branch is omitted, leaving action generation conditioned only on observed context. Appendix~\ref{app:tactile_details} provides the complete token construction, attention graph, flow-matching objectives, and inference procedure.

\section{Experiments}
\label{sec:experiments}
We organize our experiments around four questions. \textbf{First, how effective is \simpletouch{}?} We benchmark it against seven strong baselines in simulation and four on the real robot, including the pretrained tactile policy FTP-1 (Section~\ref{sec:main_results}). \textbf{Second, what makes \simpletouch{} effective?} We use controlled ablations to examine its two central principles: preserving pretrained tactile representations and learning their task-specific use through predictive supervision (Section~\ref{sec:ablations}). \textbf{Third, how data-efficient is task-level tactile learning?} We evaluate performance across different demonstration budgets to determine whether the method remains effective when task data are scarce (Section~\ref{sec:data_efficiency}). \textbf{Finally, does predictive tactile learning require additional inference-time generation?} We compare alternative ways of coupling tactile prediction with action generation and evaluate both task performance and computational efficiency (Section~\ref{sec:efficiency}).

\subsection{Experimental Setup}
\paragraph{Benchmarks and Tasks.}
Our evaluation focuses on contact-rich manipulation in both simulation and the real world. Following FTP-1~\citep{yuan2026ftp}, we select six tasks from UniVTAC~\citep{chen2026univtac}: Lift Bottle, Pull-out Key, Lift Can, Put Bottle, Insert Hole, and Insert Tube. Our real-robot suite comprises Play Mahjong, Wipe Board, Pick Up Chips, and Insert USB; representative rollouts are shown in Figure~\ref{fig:real_rollouts}. Every method is trained on the same 50 task-specific demonstrations per task. We evaluate 100 episodes per method and task on UniVTAC and 20 on the real robot.

\paragraph{Baselines and Metrics.}
On UniVTAC, we compare \simpletouch{} with seven strong baselines. ACT~\citep{zhao2023act} is a task-trained visuomotor policy and serves as a vision-only reference without policy pretraining, while \pihalf{}~\citep{intelligence2025pi05} provides a strong pretrained VLA reference. VITaL~\citep{george2024vital} first aligns visual and tactile representations; UniVTAC-ACT~\citep{chen2026univtac} augments ACT with tactile observations; and Tactile-VLA~\citep{huang2025tactilevla} injects tokenized touch into a pretrained vision-language backbone. FTP-$\pi_{0.5}$ uses the FTP-1 architecture initialized from \pihalf{} but without tactile policy pretraining, whereas FTP-1~\citep{yuan2026ftp} is pretrained on large-scale heterogeneous tactile manipulation data. The latter is therefore our principal comparison to a pretrained tactile policy. The real-robot evaluation includes ACT, \pihalf{}, UniVTAC-ACT, and FTP-1. We report the task success rate and the unweighted average across tasks. Further experimental details are provided in Appendices~\ref{app:implementation},~\ref{app:protocol}, and~\ref{app:variants}.

\begin{table}[t]
\centering
\caption{Results on the UniVTAC simulation benchmark. We report success rates (\%). ``Avg. w/o Lifts'' excludes Lift Bottle and Lift Can. We highlight the \textbf{best} and \underline{second-best} results.}
\label{tab:main}
\vspace{0.5em}
\fontsize{7.5pt}{8.5pt}\selectfont
\setlength{\tabcolsep}{0pt}
\begin{tabular}{@{}p{0.135\linewidth}*{7}{>{\centering\arraybackslash}p{0.10625\linewidth}}>{\centering\arraybackslash}p{0.12125\linewidth}@{}}
\toprule
\textbf{Method} & \textbf{Lift Bottle} & \textbf{Pull-out Key} & \textbf{Lift Can} & \textbf{Put Bottle} & \textbf{Insert Hole} & \textbf{Insert Tube} & \textbf{Avg.} & \textbf{Avg. w/o Lifts}\\
\midrule
\textbf{ACT} &42&28&20&28&19&45&30.3&30.0\\
{\boldmath$\pi_{0.5}$} &92&36&59&31&26&34&46.3&31.8\\
\textbf{VITaL} &72&47&8&32&25&34&36.3&34.5\\
\textbf{UniVTAC-ACT} &71&46&29&31&24&56&42.8&39.3\\
\textbf{Tactile-VLA} &\underline{97}&32&15&10&41&56&41.8&34.8\\
\textbf{FTP-}{\boldmath$\pi_{0.5}$} &77&30&26&19&47&72&45.2&42.0\\
\textbf{FTP-1} &\underline{97}&\underline{48}&\underline{65}&\underline{47}&\underline{64}&\underline{79}&\underline{66.7}&\underline{59.5}\\
\midrule
\textbf{\simpletouch{}} &\textbf{100}&\textbf{62}&\textbf{68}&\textbf{50}&\textbf{85}&\textbf{100}&\textbf{77.5}&\textbf{74.3}\\
\bottomrule
\end{tabular}
\vspace{-0.9em}
\end{table}

\begin{table}[t]
\centering
\caption{Results on four real-world contact-rich manipulation tasks. We report success rates (\%). We highlight the \textbf{best} and \underline{second-best} results.}
\label{tab:real}
\vspace{0.5em}
\fontsize{7.5pt}{8.5pt}\selectfont
\setlength{\tabcolsep}{0pt}
\begin{tabular}{@{}p{0.20\linewidth}*{5}{>{\centering\arraybackslash}p{0.16\linewidth}}@{}}
\toprule
\textbf{Method} & \textbf{Play Mahjong} & \textbf{Wipe Board} & \textbf{Pick Up Chips} & \textbf{Insert USB} & \textbf{Avg.}\\
\midrule
\textbf{ACT} &45&15&30&0&22.5\\
{\boldmath$\pi_{0.5}$} &40&45&55&\underline{10}&37.5\\
\textbf{UniVTAC-ACT} &70&35&25&0&32.5\\
\textbf{FTP-1} &\underline{95}&\underline{65}&\underline{60}&\textbf{30}&\underline{62.5}\\
\midrule
\textbf{\simpletouch{}} &\textbf{100}&\textbf{85}&\textbf{70}&\textbf{30}&\textbf{71.3}\\
\bottomrule
\end{tabular}
\vspace{-0.6em}
\end{table}

\FloatBarrier
\subsection{Main Results}
\label{sec:main_results}
\paragraph{Simulation Results.}
Table~\ref{tab:main} reports results on UniVTAC. With the same 50 task-specific demonstrations, \simpletouch{} ranks first on all six tasks and reaches an average success rate of 77.5\%. This lead spans both visually dominated and contact-sensitive tasks. Among the evaluated tactile policies without tactile policy pretraining, FTP-$\pi_{0.5}$ is the strongest baseline at 45.2\%; \simpletouch{} improves this average by 32.3 percentage points and leads it on every task, with gains ranging from 23 points on Lift Bottle to 42 points on Lift Can. \simpletouch{} also surpasses the pretrained FTP-1 policy on all six tasks, improving the overall average by 10.8 points. The advantage remains substantial beyond the visually dominated lift tasks: excluding Lift Bottle and Lift Can, \simpletouch{} averages 74.3\%, compared with 59.5\% for FTP-1. Its largest gains over FTP-1 occur on Insert Hole and Insert Tube, where it improves success by 21 points on each task.

\paragraph{Real-Robot Results.}
\label{sec:real}
Across the four real-world tasks in Table~\ref{tab:real}, \simpletouch{} raises average success from 62.5\% for FTP-1 to 71.3\%, an 8.8-percentage-point gain. It leads on Play Mahjong, Wipe Board, and Pick Up Chips, and ties FTP-1 on Insert USB. The 20-point gain on Wipe Board and 10-point gain on Pick Up Chips indicate benefits in sustained-contact and small-object manipulation, respectively. Insert USB remains challenging at 30\% for both methods. Together with the UniVTAC results, these experiments show that strong pretrained representations and a task-trained tactile expert can support accurate contact-rich control without tactile policy pretraining.

\begin{figure}[t]
    \centering
    \includegraphics[width=\linewidth]{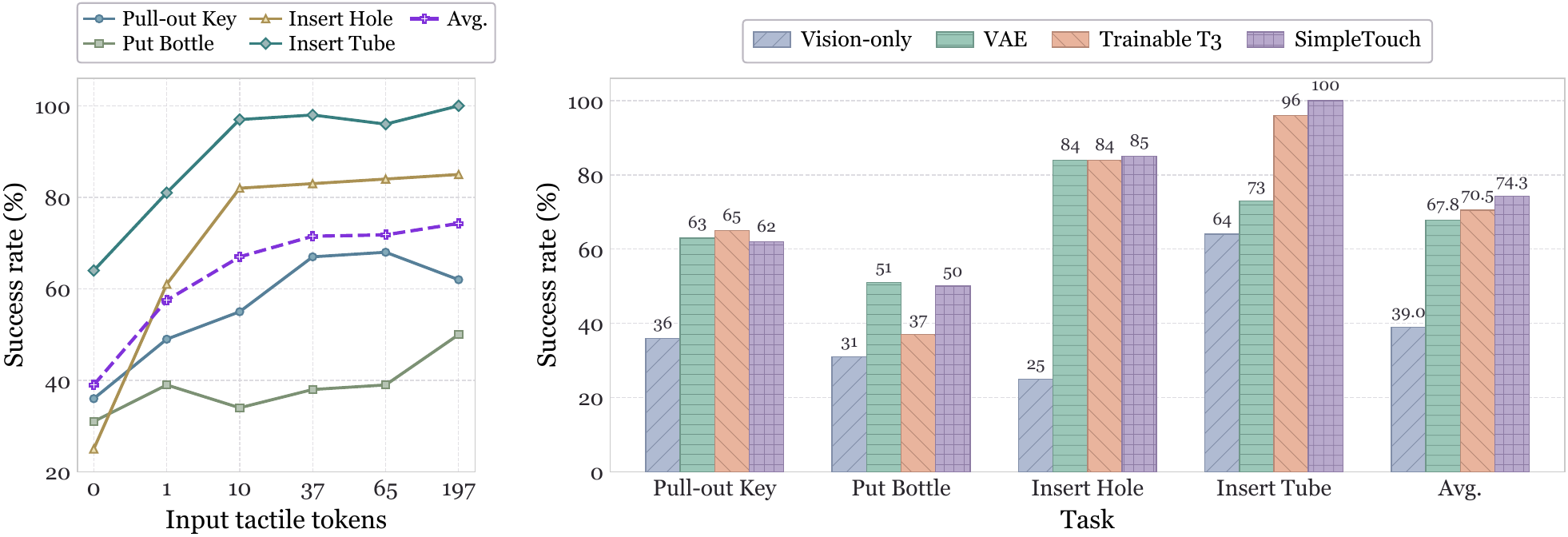}
    \caption{\textbf{\textit{(Left)}} Effect of the number of retained current tactile tokens per sensor; fewer tokens imply stronger spatial pooling, 0 is vision-only, 1 retains only the T3 CLS token, and 197 is the full T3 sequence. \textbf{\textit{(Right)}} Effects of tactile-encoder variants across tasks; VAE uses a frozen visual VAE, while Trainable T3 fine-tunes T3.}
    \label{fig:ablations}
    \vspace{1.5em}
    \includegraphics[width=\linewidth]{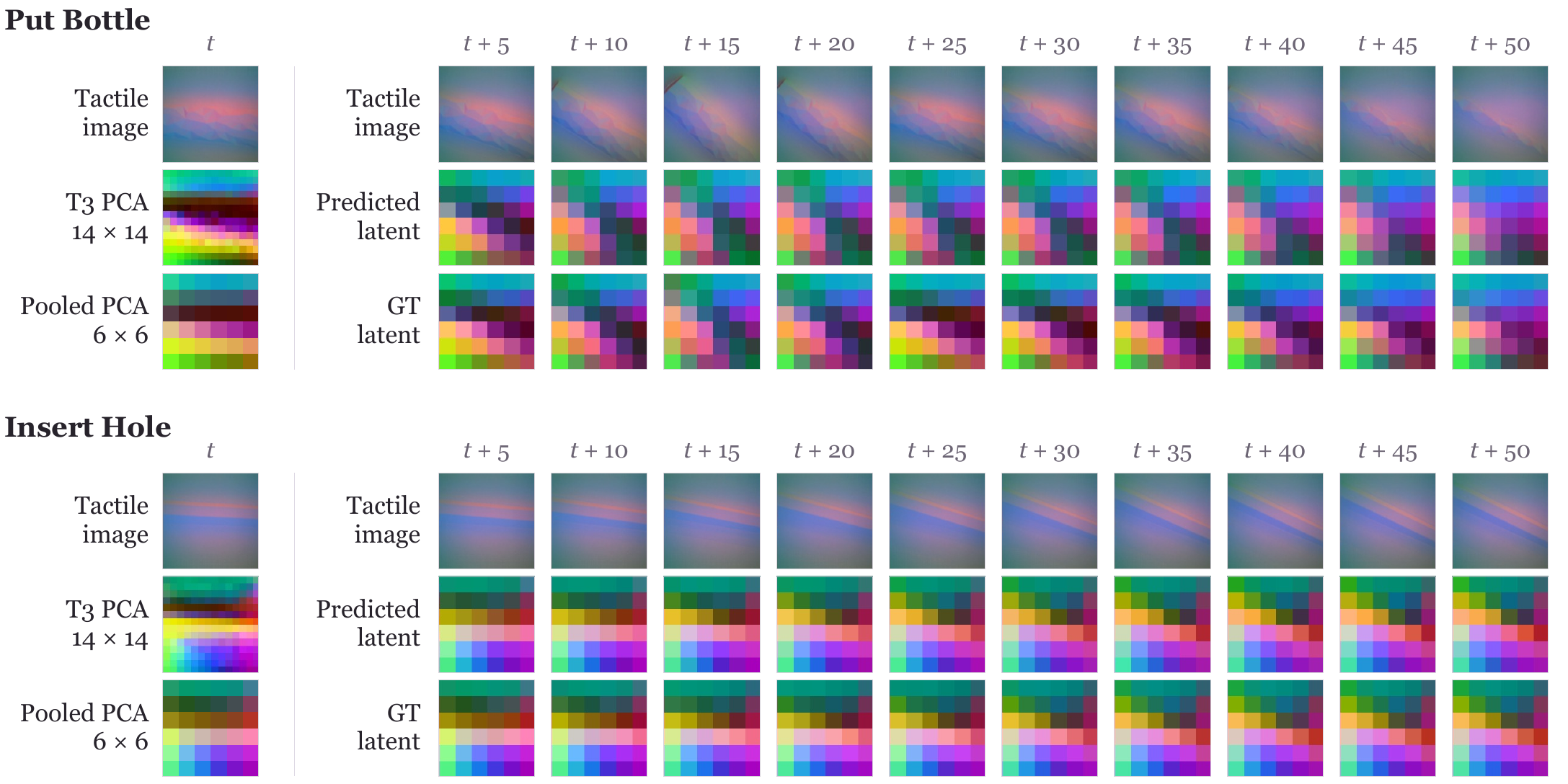}
    \caption{Multi-horizon future tactile prediction. The current column shows the tactile image, the $14\!\times\!14$ spatial T3 PCA map, and its $6\!\times\!6$ pooling. Future columns at 5-step intervals show the observed image and PCA maps of predicted and ground-truth pooled latents.}
    \label{fig:tactile_prediction}
    \vspace{-0.9em}
\end{figure}

\paragraph{Qualitative Behavior Analysis.}
Rollouts reveal a clear progression in how policies use touch across visually ambiguous and dynamically changing contact states. Vision-only policies struggle to resolve ambiguous contact states in the observed rollouts, while policies with weaker tactile integration still misread contact, slip, or apply ineffective corrections. FTP-1 often performs deliberate probing before committing to an action, consistent with contact-acquisition skills encouraged by large-scale tactile policy pretraining; this is particularly valuable for fine-grained manipulation under uncertain contact. \simpletouch{} shows no equally explicit probing phase, but interprets the available tactile feedback accurately and makes decisive corrections while maintaining stable grasp, contact, and insertion.

\FloatBarrier
\subsection{Ablation Studies}
\label{sec:ablations}
\paragraph{Tactile Representation.}
Figure~\ref{fig:ablations} examines the representation supplied to the tactile expert. Spatial pooling controls the retained token count: fewer tokens reduce spatial resolution, whereas 197 preserves the full T3 sequence. The four-task average rises from 39.0\% without touch to 57.5\% with only the T3 CLS token and to 74.3\% with all 197 tokens. Thus, coarse touch is already useful, but spatial detail brings further gains, especially for Insert Hole, Insert Tube, and Put Bottle. The encoder comparison tests whether the tactile-expert interface remains effective with a different representation family. A pretrained visual VAE reaches 67.8\% and remains competitive on three tasks, showing that the interface can exploit a generic visual latent space; its larger drop on Insert Tube (73\% versus 100\%) shows that demanding contact tasks still benefit from tactile-specialized features. The trainable-T3 variant achieves a mean success rate of 70.5\%, below the 74.3\% obtained when its pretrained space is preserved, consistent with limited task data partially distorting transferable tactile features.

\paragraph{Future Tactile Prediction.}
\begin{wraptable}{r}{0.55\textwidth}
    \centering
    \vspace{-0.8\baselineskip}
    \caption{Future tactile prediction ablations on UniVTAC. We highlight the \textbf{best} and \underline{second-best} values.}
    \label{tab:predictive_ablation}
    \vspace{0.4em}
    \fontsize{8.5pt}{9.5pt}\selectfont
    \setlength{\tabcolsep}{0.5pt}
    \begin{tabular*}{\linewidth}{@{\extracolsep{\fill}}lccccc@{}}
    \toprule
    \textbf{Variant} & \textbf{Pull-out} & \textbf{Put} & \textbf{Insert} & \textbf{Insert} & \textbf{Avg.}\\[-0.15em]
    & \textbf{Key} & \textbf{Bottle} & \textbf{Hole} & \textbf{Tube} & \\
    \midrule
    \textbf{Vision-only} &36&31&25&64&39.0\\
    \textbf{w/o Tac. Pred.} &\underline{60}&37&\underline{83}&95&\underline{68.8}\\
    \textbf{Single-horizon} &53&\underline{44}&79&\underline{98}&68.5\\
    \midrule
    \textbf{\simpletouch{}} &\textbf{62}&\textbf{50}&\textbf{85}&\textbf{100}&\textbf{74.3}\\
    \bottomrule
    \end{tabular*}
    \vspace{-0.6\baselineskip}
\end{wraptable}

Figure~\ref{fig:tactile_prediction} compares observed tactile images with PCA projections of predicted and ground-truth pooled T3 latents. Across both tasks, the predictions qualitatively follow the changing contact structure over multiple future horizons, providing evidence that the tactile expert learns structured tactile evolution rather than an arbitrary auxiliary target. Table~\ref{tab:predictive_ablation} compares predictive supervision under the same current-touch representation and action pathway. Multi-horizon prediction raises the four-task average from 68.8\% without prediction to 74.3\%, a 5.5-point gain, and performs best on every task. Predicting the same pooled representation only at $t{+}50$ reaches 68.5\%, compared with 74.3\% for ten future horizons. With spatial resolution held fixed, this 5.8-point gain supports supervising intermediate contact evolution rather than only the final future state. The largest gain over no prediction is on Put Bottle (50\% versus 37\%), where contact changes continuously during placement; Insert Hole and Insert Tube show smaller but consistent gains after contact. Together, the qualitative fidelity and policy improvements support multi-horizon prediction as supervision for the tactile-action interface.

\FloatBarrier
\subsection{Data Efficiency}
\label{sec:data_efficiency}
To test whether \simpletouch{} remains effective when task data are scarce, we compare it with FTP-$\pi_{0.5}$ on Put Bottle and Insert Hole across four demonstration budgets. Figure~\ref{fig:data_efficiency} shows a consistent advantage at every budget. With only 10 demonstrations, \simpletouch{} reaches 23\% and 43\%, versus 5\% and 14\% for FTP-$\pi_{0.5}$. With 20 demonstrations---60\% fewer than the baseline's 50-demonstration setting---it reaches 38\% and 57\%, exceeding the corresponding baseline results of 19\% and 47\%. The advantage therefore emerges well before the standard 50-demonstration setting. Performance is not universally monotonic at larger budgets, so these results support consistent data-efficiency gains rather than a general scaling law.

\noindent
\begin{minipage}[t]{0.49\textwidth}
\vspace{0pt}
\centering
\includegraphics[width=\linewidth]{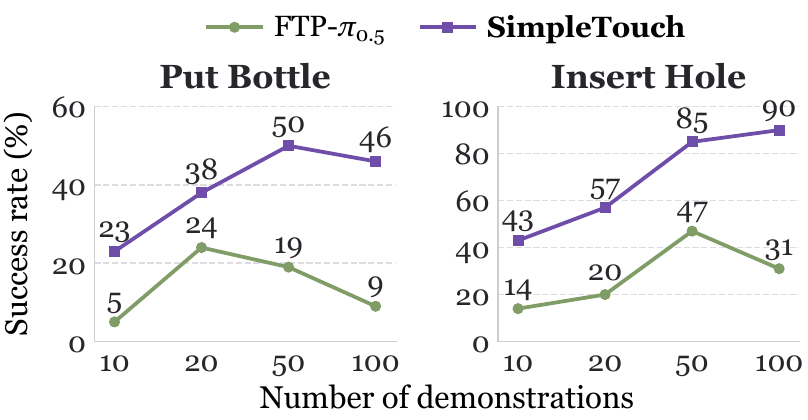}
\vspace{-1.8em}
\captionof{figure}{Data efficiency on UniVTAC. Success rates of FTP-$\pi_{0.5}$ and \simpletouch{} on Put Bottle and Insert Hole across demonstration budgets.}
\label{fig:data_efficiency}
\end{minipage}\hfill
\begin{minipage}[t]{0.49\textwidth}
\vspace{0pt}
\centering
\captionof{table}{Tactile design choices in FTP-$\pi_{0.5}$ and \simpletouch{}; shaded rows mark design choices that differ between the two methods.}
\label{tab:data_efficiency_arch}
\vspace{0.45em}
\small
\renewcommand{\arraystretch}{1.15}
\setlength{\tabcolsep}{2.2pt}
\begin{tabular}{@{}>{\raggedright\arraybackslash}p{0.45\linewidth}>{\centering\arraybackslash}p{0.23\linewidth}>{\centering\arraybackslash}p{0.26\linewidth}@{}}
\toprule
\textbf{Design choice} & \textbf{FTP-$\pi_{0.5}$} & \textbf{\simpletouch{}} \\
\midrule
Tactile encoder & T3 & T3 \\
\rowcolor{blue!7}
Encoder update & Fine-tuned & Frozen \\
\rowcolor{blue!7}
Current tactile tokens & CLS only & All tokens \\
Tactile expert & Yes & Yes \\
\rowcolor{blue!7}
Future tactile prediction & No & Multi-horizon \\
\bottomrule
\end{tabular}
\end{minipage}
\par\vspace{0.7em}

\begin{figure}[t!]
    \centering
    \includegraphics[width=\linewidth]{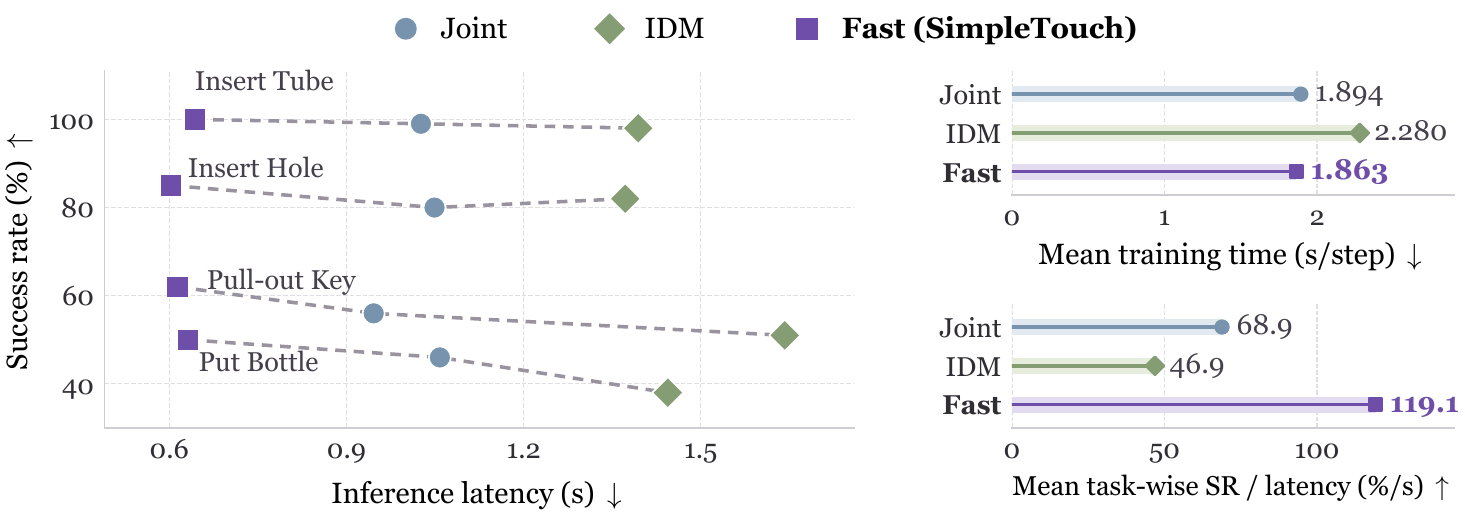}
    \mainfigcaptionspace
    \caption{Training and inference efficiency of Joint, IDM, and Fast. The left panel shows success rate versus inference latency; dashed lines connect formulations evaluated on the same task. The right panels show mean training time (top) and mean task-wise success-rate/latency ratio (bottom).}
    \label{fig:efficiency}
    \vspace{-1.2em}
\end{figure}
\FloatBarrier

Table~\ref{tab:data_efficiency_arch} helps interpret this gap. Both methods use T3 and a tactile expert, focusing the comparison on how tactile representations are preserved and supervised rather than whether touch is present. The ablations in Section~\ref{sec:ablations} support each distinction: retaining all tactile tokens preserves spatial contact detail, keeping the pretrained encoder fixed performs better than updating it in this limited-data setting, and multi-horizon prediction provides dense supervision for learning contact dynamics. Together, these choices help explain why \simpletouch{} remains effective across demonstration budgets, including the lowest-data settings.

\subsection{Training and Inference Efficiency}
\label{sec:efficiency}
We evaluate three formulations adapted from Fast-WAM~\citep{yuan2026fastwam}. All use future tactile prediction as training supervision but differ in how future latents participate in action generation: \emph{Joint} denoises tactile latents and actions together, \emph{IDM} generates future latents before predicting actions, and \emph{Fast}, used by \simpletouch{}, predicts actions directly from cached current context. Figure~\ref{fig:efficiency} summarizes their success rates and timings on the four ablation tasks.

Fast establishes the strongest accuracy--efficiency trade-off. It attains lower latency and higher success on every task, rather than improving the average through one favorable case. Its mean success rate is 74.3\%, versus 70.3\% for Joint and 67.3\% for IDM, while mean latency falls to 0.623\,s from 1.020\,s and 1.463\,s---reductions of 38.9\% and 57.4\%. This consistent upper-left shift suggests that feeding generated future latents into the action pathway is unnecessary and may introduce an avoidable dependency on prediction quality. Future-touch prediction is more effective here as a training signal for the tactile expert, while deployment relies on observed tactile context.

The conclusion holds when training cost is considered. Fast and Joint require 1.863\,s and 1.894\,s per measured training operation, a difference of only 1.6\%, whereas IDM requires 2.280\,s. Combining task success with latency, Fast reaches 119.1\,\%/s, compared with 68.9 for Joint and 46.9 for IDM---1.73 and 2.54 times as high, respectively. Fast therefore preserves predictive supervision without paying for future tactile generation at deployment, motivating its use as the default formulation in \simpletouch{}. Timing definitions, hardware, and protocols are provided in Appendix~\ref{app:efficiency_protocol}.

\section{Conclusion}
We present \simpletouch{}, a simple and effective recipe for tactile-enhanced vision-language-action policies without additional tactile policy pretraining or separate alignment. It augments \pihalf{} with a tactile expert that consumes the full token sequence of a pretrained tactile encoder, learns through action supervision and multi-horizon future-touch prediction, and acts from current tactile context at inference. With 50 demonstrations per task, \simpletouch{} leads all six UniVTAC tasks (77.5\% average) and reaches 71.3\% across four real-robot tasks, outperforming tactile-policy-pretrained FTP-1 by 10.8 and 8.8 percentage points, respectively. Ablations support preserved spatial detail, stable pretrained representations, and predictive supervision as useful ingredients, providing a practical and efficient route to contact-rich manipulation.

\section*{AI Use Statement}
Generative AI tools were used to assist with literature retrieval, research ideation and experimental planning, software implementation, result analysis and visualization, and the drafting and polishing of this manuscript. All AI-assisted code, citations, analyses, figures, and text were reviewed and verified by the authors. The authors take full responsibility for the content and claims of this work.

\section*{Ethics Statement}
This work studies tactile-enhanced robot manipulation in simulation and on real robotic platforms. The experiments involve no human subjects, personal data, or deployment in safety-critical environments, and all real-robot evaluations were conducted under controlled laboratory conditions. We do not identify direct ethical risks beyond the standard safety considerations associated with robotic manipulation.

\section*{Reproducibility Statement}
We support reproducibility by describing the model and learning objectives in Section~3, the evaluation protocol and principal experimental settings in Section~4, and additional implementation, training, hardware, and evaluation details in the appendices. We are preparing a public release of the complete codebase and model checkpoints, together with all training configurations and evaluation scripts, and will make these resources available in the near future.

\bibliography{references}
\bibliographystyle{iclr2027_conference}
\clearpage
\appendix
\newcommand{\appmissing}[2]{\par\smallskip\noindent\colorbox{yellow!35}{\parbox{\dimexpr\linewidth-2\fboxsep\relax}{\small\textbf{#1 (pending).} #2}}\par\smallskip}
\newcommand{\appimage}[2]{\noindent\fcolorbox{black!25}{yellow!20}{\parbox[c][1.05in][c]{\dimexpr\linewidth-2\fboxsep-2\fboxrule\relax}{\centering\textbf{#1}\par\smallskip\small #2}}}
\newcommand{\apptablecaptionspace}{\vspace{0.15em}}

\section{Implementation Details}
\label{app:implementation}
\subsection{Model, Training, and Inference Configuration}

Unless otherwise specified, all \simpletouch{} experiments use the model, training, and inference configuration summarized in Table~\ref{tab:config}. The values report the resolved default configuration used by the training launcher rather than inactive model defaults.

\begin{center}
\captionof{table}{Model, training, and inference configurations for \simpletouch{}.}
\label{tab:config}
\vspace{-0.9em}
\footnotesize
\renewcommand{\arraystretch}{0.96}
\setlength{\tabcolsep}{6pt}
\begin{tabular}{>{\raggedright\arraybackslash}p{0.39\linewidth}>{\raggedright\arraybackslash}p{0.55\linewidth}}
\toprule
Configuration & Value \\
\midrule
\multicolumn{2}{l}{\textit{Vision-language-action backbone}}\\
Base policy & $\pi_{0.5}$ \\
Visual-language backbone & PaliGemma~\citep{beyer2024paligemma}: SigLIP~\citep{zhai2023siglip} vision encoder and Gemma 2B language model \\
Action expert & Gemma 300M; width 1024; 18 Transformer layers~\citep{vaswani2017attention} \\
State and action spaces & Shared 8-dimensional interface: UniVTAC uses seven arm joints and one gripper coordinate; PiPER uses six arm joints, one zero dummy joint, and one gripper coordinate. States are absolute; action-joint dimensions are relative and the gripper action is absolute. Both are zero-padded to 32 internal model dimensions \\
Action chunk length & 50 \\
\midrule
\multicolumn{2}{l}{\textit{Tactile representation and expert}}\\
Tactile encoder & Pretrained T3-large; frozen during task training \\
Tactile preprocessing & Raw RGB; resize, center crop to $224\times224$, and task-specific normalization \\
Current tactile tokens & 2 sensors $\times$ (1 CLS + $14\times14$ spatial tokens) = 394 tokens; feature dimension 1024 \\
Tactile expert & Independently initialized; same width and depth as the action expert \\
Prediction formulation & Fast; actions attend to current tactile features but not future tactile targets \\
Future horizons & $+5,+10,\ldots,+50$ \\
Future target tokens & 2 sensors $\times$ (1 CLS + $6\times6$ pooled spatial tokens) per horizon; 740 tokens in total \\
Auxiliary objective & Flow-matching velocity MSE; $\lambda_T=0.01$ with a constant schedule \\
Optimization scope & Action and tactile objectives jointly update the VLA and tactile expert \\
Episode-tail supervision & Repeat the final valid action and tactile target and retain supervision \\
\midrule
\multicolumn{2}{l}{\textit{Training configuration}}\\
Task data & 50 task-specific demonstrations per task \\
Training duration & 20,000 updates; checkpoint every 5,000 updates \\
Optimizer & AdamW~\citep{loshchilov2017adamw}; $\beta_1=0.9$, $\beta_2=0.95$, $\epsilon=10^{-8}$ \\
Learning rate & Peak $5\times10^{-5}$; 1,000-step warmup; cosine decay to $5\times10^{-6}$ \\
Weight decay / gradient clipping & $10^{-10}$ / global norm 1.0 \\
Batch size & 8 samples per GPU $\times$ 8 GPUs = 64 \\
Training hardware & 8 NVIDIA H20 GPUs \\
\midrule
\multicolumn{2}{l}{\textit{Inference configuration}}\\
Flow solver & 10 Euler integration steps \\
Prediction / execution horizon & Predict 50 actions; execute 25 for Insert Hole and Insert Tube and 50 for the other tasks \\
Tactile computation & Encode current touch once per policy query and reuse its layerwise KV cache during action denoising \\
Additional inference operations & No future tactile generation or temporal ensembling \\
\bottomrule
\end{tabular}
\end{center}

\subsection{Tactile Expert Training and Inference}
\label{app:tactile_details}

This section gives the exact token construction, conditioning graph, and optimization objective used by the Fast formulation of \simpletouch{}. It also makes explicit which computations are retained at deployment. These details complement the architectural description in Section~\ref{sec:method}; they do not introduce an additional training stage.

\paragraph{Current and Future Tactile Tokens.}
Let $E_\phi$ denote the pretrained T3 encoder and let $U_t^l$ and $U_t^r$ be the synchronized observations from the two tactile sensors. For current touch, we concatenate the complete output of both sensors,
\begin{equation}
    Z_t=\big[E_\phi(U_t^l);E_\phi(U_t^r)\big]
    \in\mathbb{R}^{394\times1024}.
    \label{eq:app_current_tactile}
\end{equation}
Each sensor contributes one global token and a $14\times14$ spatial grid. No pooling is applied to $Z_t$, and a learned sensor-identity embedding distinguishes the two streams. Future prediction uses the offsets
$\Delta=\{5,10,\ldots,50\}$. For each sensor and offset, the global token is retained and the $14\times14$ spatial grid is pooled to $6\times6$. The resulting target is
\begin{equation}
    Y_t=
    \operatorname*{Stack}_{\delta\in\Delta}
    \operatorname*{Stack}_{s\in\{l,r\}}
    \mathcal S\!\left(\mathcal P(E_\phi(U_{t+\delta}^{s}))\right)
    \in\mathbb{R}^{10\times2\times37\times1024},
    \label{eq:app_future_tactile}
\end{equation}
where $\mathcal P$ denotes spatial pooling and $\mathcal S(x)=x/r$ uses the checkpoint-derived reference
\begin{equation}
    r=\sqrt{\frac{1}{D}\sum_{d=1}^{D}\left(\gamma_d^2+\beta_d^2\right)},
    \label{eq:app_reference_rms}
\end{equation}
with $\gamma,\beta\in\mathbb R^D$ from the final LayerNorm of the pretrained T3 encoder. This fixed scaling is applied only to future targets; the current tactile context in Eq.~\eqref{eq:app_current_tactile} remains unmodified. The horizon and sensor axes are flattened into 740 future tokens before the tactile expert. Near the end of a trajectory, the final valid action and tactile observation are repeated so that all horizons retain supervision.

\paragraph{Independent Flow Paths.}
We train the action chunk $A_t$ and future tactile target $Y_t$ with two conditional flow-matching paths. For $q\in\{A,T\}$, let $x_q$ denote the corresponding clean target, $\epsilon_q\sim\mathcal N(0,I)$, and $\tau_q\sim\operatorname{Beta}(1.5,1)$. The action and tactile times are sampled independently and mapped to $[0.001,1]$. Their interpolants and velocity targets are
\begin{equation}
    x_q^{\tau_q}=\tau_q\epsilon_q+(1-\tau_q)x_q,
    \qquad
    v_q^\star=\epsilon_q-x_q.
    \label{eq:app_flow_paths}
\end{equation}
The two losses are
\begin{align}
    \mathcal L_{\mathrm{act}}
    &=\mathbb E\!\left[
      \operatorname{MSE}\!\left(
      f_A(x_A^{\tau_A},\tau_A;C_t,Z_t),v_A^\star\right)
      \right],
      \label{eq:app_action_loss}\\
    \mathcal L_{\mathrm{tac}}
    &=\mathbb E\!\left[
      \operatorname{MSE}\!\left(
      f_T(x_T^{\tau_T},\tau_T;C_t,Z_t),v_T^\star\right)
      \right],
      \label{eq:app_tactile_loss}
\end{align}
where $C_t$ is the visual-language and state context and $\operatorname{MSE}$ averages the squared error over the complete velocity tensor. Thus, the tactile expert reads the full noised future sequence, including all horizons, while every equally sized horizon receives equal weight. The final objective is
\begin{equation}
    \mathcal L=\mathcal L_{\mathrm{act}}+\lambda_T\mathcal L_{\mathrm{tac}},
    \qquad \lambda_T=0.01.
    \label{eq:app_total_loss}
\end{equation}
The action and tactile branches share the same visual-language context and current-touch cache, but not their noise or flow time. Consequently, tactile prediction supplies an additional learning signal without forcing action denoising to follow the prediction trajectory.

\paragraph{Block Attention Structure.}
The Fast formulation uses a fixed asymmetric visibility pattern. Rows below denote query blocks and columns denote the key-value blocks they may read:
\begin{equation}
\begin{array}{c|cccc}
\text{query}\backslash\text{key/value}
    & C_t & Z_t & \widetilde{Y}_t & \widetilde{A}_t \\
\hline
C_t                 & \checkmark & -          & -          & - \\
Z_t                 & -          & \checkmark & -          & - \\
\widetilde{Y}_t     & \checkmark & \checkmark & \checkmark & - \\
\widetilde{A}_t     & \checkmark & \checkmark & -          & \checkmark
\end{array}
\label{eq:app_attention_graph}
\end{equation}
Here $\widetilde{Y}_t$ and $\widetilde{A}_t$ are the noised future-tactile and action tokens. The four blocks correspond to $c_0$, $z_0$, $z_{1:H}$, and $a_{1:H}$ in Figure~\ref{fig:method_overview}; its future-tactile labels schematically span the sampled offsets $\Delta$, not every action step. Current tactile queries are deliberately isolated from the visual-language stream, preserving a modality-specific tactile computation. Future-tactile queries can use the current multimodal context to model contact evolution. Action queries read current touch but never the future-tactile branch; conversely, future-tactile queries never read action tokens. Thus, future tactile observations act as prediction targets rather than privileged action inputs.

\paragraph{Fast Inference and Cache Reuse.}
At each policy query, the visual-language backbone first produces the layerwise context cache $\mathcal K_C$. The tactile expert processes $Z_t$ once using tactile self-attention with the fixed clean-time condition $\tau=0$ and produces $\mathcal K_T=\mathcal K_t$ from Eq.~\eqref{eq:tactile_cache}; we suppress the environment-step index $t$ on cache components below. At every Transformer layer, their keys and values are separately concatenated along the sequence dimension,
\begin{equation}
    \mathcal K_{C,T}^{j}
    =\left(
    \operatorname{Concat}_{\mathrm{seq}}(K_C^j,K_T^j),
    \operatorname{Concat}_{\mathrm{seq}}(V_C^j,V_T^j)
    \right),
    \qquad j=1,\ldots,L.
    \label{eq:app_cache_composition}
\end{equation}
The action solver samples $x_A^0\sim\mathcal N(0,I)$ and performs $N=10$ Euler steps,
\begin{equation}
    x_A^{k+1}=x_A^k-\frac{1}{N}
    f_A\!\left(x_A^k,1-\frac{k}{N};\mathcal K_{C,T}\right),
    \qquad k=0,\ldots,N-1.
    \label{eq:app_fast_solver}
\end{equation}
The clean context cache is reused at every step because neither $C_t$ nor $Z_t$ depends on action noise or flow time. The future-tactile branch, its noisy tokens, and its output projection are omitted at inference. Multi-horizon prediction therefore trains the same tactile expert that constructs $\mathcal K_T$, while deployment retains only current-touch encoding and action generation.

\section{Extended Related Work}
\label{app:extended_related_work}

This section expands the three research directions summarized in Section~\ref{sec:related}, retaining the same organization and distinction between policy learning and tactile representation learning.

\subsection{Generalist Policy Learning}
Task-specific imitation learning established strong sequence-modeling and generative-control baselines through action chunking and diffusion~\citep{zhao2023act,chi2023diffusionpolicy}. Generalist policy learning broadens this setting by combining heterogeneous robot data with pretrained vision-language representations. Open X-Embodiment and DROID provide large robot-interaction corpora~\citep{oneill2023openx,khazatsky2024droid}, while OpenVLA, the $\pi$ family, GR00T N1, and related systems develop reusable policies across tasks and embodiments~\citep{kim2025openvla,black2024pi0,intelligence2025pi05,bjorck2025groot}. FAST action tokenization and OFT further study efficient action representations and task fine-tuning~\citep{pertsch2025fast,kim2025oft}. These systems primarily organize vision, language, proprioception, and actions; local tactile contact is generally not native to their pretrained observation interfaces.

A complementary world-action-model line incorporates predictive visual or latent dynamics into control. Video Prediction Policy predicts future visual features~\citep{hu2024vpp}, whereas Motus, DreamZero, LingBot-VA, LaWAM, and Fast-WAM couple visual or latent prediction with action generation in different training and inference formulations~\citep{bi2025motus,ye2026dreamzero,li2026lingbot,zhang2026lingbot2,chen2026lawam,yuan2026fastwam}. Collectively, these works motivate prediction as supervision beyond behavioral cloning, but they still predominantly model visual rather than tactile dynamics.

\subsection{Tactile Policy Learning}
As in Section~\ref{sec:related}, we organize tactile policies by where their contact-sensitive priors are acquired. \emph{Tactile policy pretraining and mid-training} learn tactile control before downstream adaptation: FTP-1 and the $N_0$ family scale tactile-native policy learning, T-Rex introduces tactile-grounded mid-training and fast reaction, and TouchWorld combines tactile prediction with hierarchical control~\citep{yuan2026ftp,neote2026n0twam,neote2026n0vtla,niu2026trex,zhou2026touchworld}. \emph{Representation alignment} instead reduces the modality gap before policy learning through shared, sequential, or semantically aligned visual-tactile spaces, as explored by VITaL, UniTouch, VT-MUSE, and OmniVTLA~\citep{george2024vital,yang2024unitouch,xu2026vtmuse,cheng2025omnivtla}.

\emph{Task-level tactile integration} directly equips pretrained policies with contact observations. Tactile-VLA, AT-VLA, and TacFiLM respectively emphasize tactile reasoning, adaptive injection, and feature-wise modulation, while ForceVLA and FM-VLA use force-aware expert routing or force-history memory~\citep{huang2025tactilevla,li2026atvla,morissette2026tacfilm,yu2025forcevla,li2026fmvla}. Other systems similarly study dual-level feedback, structured visuotactile features, or fast tactile reaction~\citep{bi2025vlatouch,huang2024threedvitac,heng2025vitacformer,xue2025rdp}.

\emph{Predictive tactile learning} models future contact as supervision or a control signal. Dream-Tac, DreamTacVLA, Representation-Aligned Tactile Grounding, UniTacVLA, Touch Dreaming, and OmniVTA span visuotactile world models, future-touch prediction, representation-level grounding, and tactile-conditioned correction~\citep{lou2026dreamtac,ye2025dreamtacvla,chen2026representationgrounding,zhang2026unitacvla,niu2026touchdreaming,zheng2026omnivta}. Although these categories can overlap, they distinguish whether additional tactile knowledge comes from policy pretraining, representation alignment, task-level fusion, or predictive supervision.

\subsection{Pretrained Tactile Encoders}
Pretrained tactile encoders address sensory representation acquisition rather than action generation. T-Dex learns self-supervised embeddings from robotic play~\citep{guzey2023tdex}; Sparsh and Sparsh-X study self-supervised visual-tactile and multisensory representations~\citep{higuera2024sparsh,higuera2025sparshx}; and UniTouch and the AnyTouch series target transfer across sensors, modalities, and temporal tactile signals~\citep{yang2024unitouch,feng2025anytouch,feng2026anytouch2}. T3 combines sensor-specific input processing with a shared Transformer trained across diverse sensors and tasks~\citep{zhao2025t3}. Masked visual-tactile modeling, tactile dexterity benchmarks, and large-scale touch-centric datasets provide complementary objectives and resources for reusable tactile features~\citep{liu2024maskedvt,liu2025vtdex,cheng2025touch100k}.

Data collection systems and simulators also shape the resulting representations. Robot-free interfaces such as ViTaMIn and FreeTacMan expand tactile interaction data without conventional robot teleoperation~\citep{liu2025vitamin,wu2025freetacman}, while UniVTAC supplies controlled visuotactile trajectories and standardized evaluation~\citep{chen2026univtac}. Such encoders provide transferable contact features, but do not prescribe how those features should influence a pretrained action generator. This separates tactile representation pretraining from tactile policy pretraining and motivates studying the downstream policy interface independently.

\section{Benchmarks and Evaluation Protocols}
\label{app:protocol}
\subsection{UniVTAC Tasks and Demonstrations}
The main simulation study follows the six-task UniVTAC subset used by FTP-1~\citep{chen2026univtac,yuan2026ftp}. All evaluated methods use the same 50 official task-specific demonstrations and are evaluated for 100 episodes per task. Each task has a separately fine-tuned policy, and no trajectories are shared across tasks.

\paragraph{Action Prediction and Execution Horizons.}
Table~\ref{tab:app_simulation_protocol} separates the 50-step action chunk predicted at each policy query from the number of actions executed before requesting a new observation. We specify the execution horizon before evaluation from task characteristics rather than tuning it on validation or test performance. The non-insertion tasks execute the full predicted chunk, whereas Insert Hole and Insert Tube execute only the first 25 actions. These insertion tasks require repeated contact-conditioned corrections while the object is aligned with a constrained opening. The shorter execution horizon refreshes the tactile observation midway through the predicted chunk, allowing the policy to reassess both the alignment error and the appropriate correction direction after contact changes. Executing a longer open-loop prefix can instead preserve a stale correction after the contact geometry has changed, leading to overshoot, jamming, slippage, or the object moving away from the insertion region. All tasks retain the same prediction horizon and ten action-flow integration steps; only the feedback frequency differs. Baseline results use the official default execution settings of their respective methods.

\begin{center}
\centering
\captionof{table}{UniVTAC task protocol for \simpletouch{}. ``Predicted'' and ``Executed'' denote the action-prediction and execution horizons, respectively.}
\label{tab:app_simulation_protocol}
\apptablecaptionspace
\small
\setlength{\tabcolsep}{6pt}
\begin{tabular}{lccccc}
\toprule
Task & Demonstrations & RGB views & Predicted & Executed & Eval. trials\\
\midrule
Lift Bottle & 50 & Head & 50 & 50 & 100\\
Pull-out Key & 50 & Head & 50 & 50 & 100\\
Lift Can & 50 & Head + wrist & 50 & 50 & 100\\
Put Bottle & 50 & Head & 50 & 50 & 100\\
Insert Hole & 50 & Head & 50 & 25 & 100\\
Insert Tube & 50 & Head + wrist & 50 & 25 & 100\\
\bottomrule
\end{tabular}
\end{center}

\paragraph{Task Selection.}
The complete UniVTAC benchmark contains eight tasks spanning pose reasoning, shape perception, and contact-rich interaction. Following FTP-1, we omit Grasp Classify and Insert HDMI. Grasp Classify is already saturated by prior tactile baselines, which reach 99--100\% success, and therefore provides little resolution for comparing tactile policies. The official Insert HDMI demonstrations are generated by motion planning and complete insertion perfectly, leaving insufficient corrective contact feedback for evaluating whether a policy learns tactile-dependent adjustment. The remaining six tasks cover in-hand and pose-sensitive manipulation (Lift Bottle, Lift Can, and Put Bottle) together with contact-aware insertion and extraction (Pull-out Key, Insert Hole, and Insert Tube)~\citep{yuan2026ftp}.

\begin{figure}[htbp]
\centering
\includegraphics[width=\linewidth]{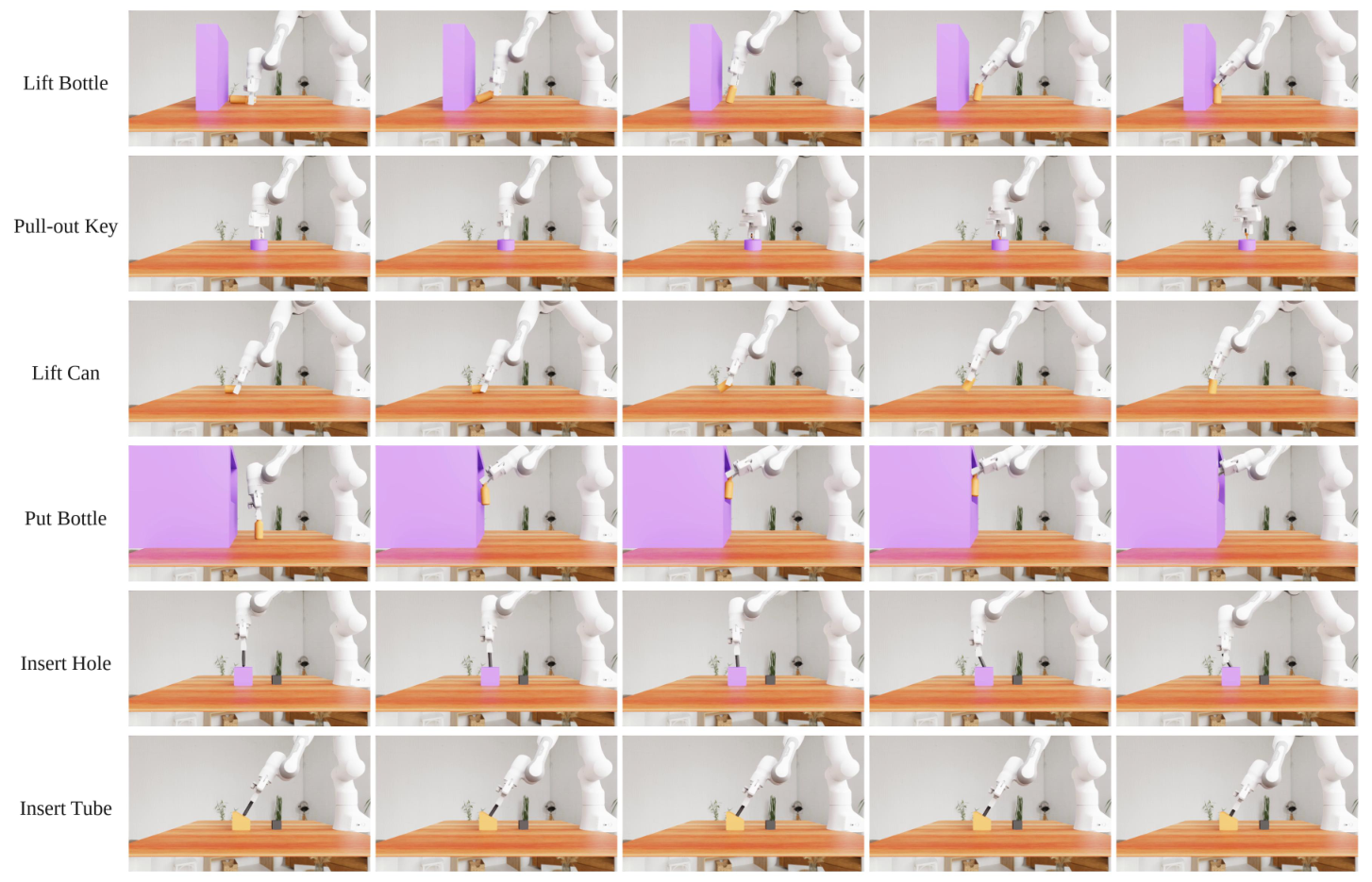}
\caption{UniVTAC simulation task rollouts. Each row corresponds to one of the six evaluated tasks and shows five key frames illustrating its manipulation and contact progression~\citep{chen2026univtac,yuan2026ftp}.}
\label{fig:app_univtac_rollouts}
\end{figure}

\paragraph{Task Definitions and Data.}
We run the official UniVTAC environments in NVIDIA Isaac Sim 4.5.0 with TacEx 0.1.0, retaining the benchmark's task randomization, demonstrations, and success definitions~\citep{chen2026univtac}. In Lift Bottle, the policy grasps and raises a bottle while keeping its base within 5\,cm of a nearby wall. Pull-out Key randomizes the key's initial rotation and requires the policy to rotate against mechanical resistance before extracting it. Lift Can samples among cylindrical cans of 4, 5, and 6\,cm diameter and requires a stable lift without slippage. Put Bottle requires grasping an upright bottle and placing it inside a shelf cavity. Insert Hole presents an inclined hole at either $60^\circ$ or $120^\circ$; the policy must infer its orientation through contact and insert a test tube. Insert Tube requires inserting a 2.0\,cm tube into a 2.05\,cm opening on an inclined surface. The training set contains 50 automatically collected full trajectories per task, and evaluation uses 100 rollouts with the benchmark's task-specific completion predicates. As in UniVTAC, excessive gel penetration or substantial relative slip violates the physics-based validity conditions rather than constituting successful completion. Figure~\ref{fig:app_univtac_rollouts} visualizes the six tasks under this protocol.

\subsection{Real-Robot Setup and Task Definitions}
The real-robot study covers Play Mahjong, Wipe Board, Pick Up Chips, and Insert USB. These tasks require tactile discrimination, contact-state estimation, force-sensitive grasping, and precise insertion, respectively. Every method uses 50 task-specific demonstrations and is evaluated for 20 rollouts per task. We reset the arm to a recorded task-specific configuration before each rollout and randomize task-relevant object conditions within the demonstrated range.

\begin{figure}[htbp]
\centering
\includegraphics[width=\linewidth]{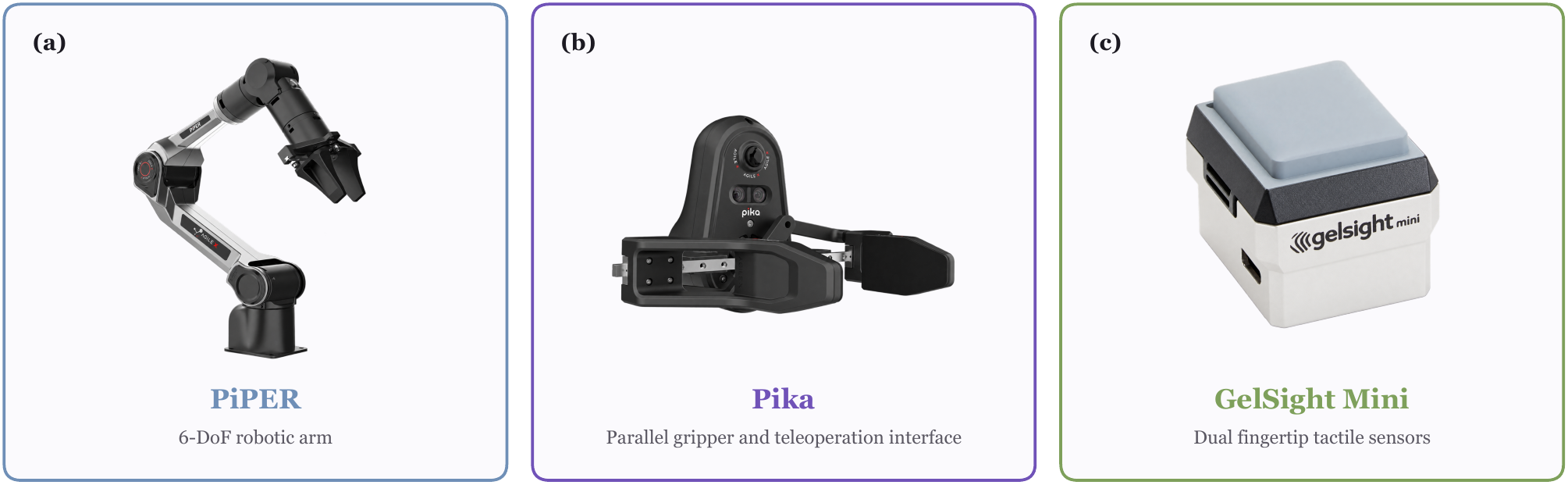}
\caption{Hardware used for real-robot experiments. (a) PiPER six-degree-of-freedom arm. (b) Pika parallel gripper and teleoperation interface. (c) GelSight Mini tactile sensor; two units are mounted on opposing gripper fingers. Product images are from the respective manufacturers.}
\label{fig:app_hardware}
\end{figure}

\begin{figure}[htbp]
\centering
\includegraphics[width=\textwidth]{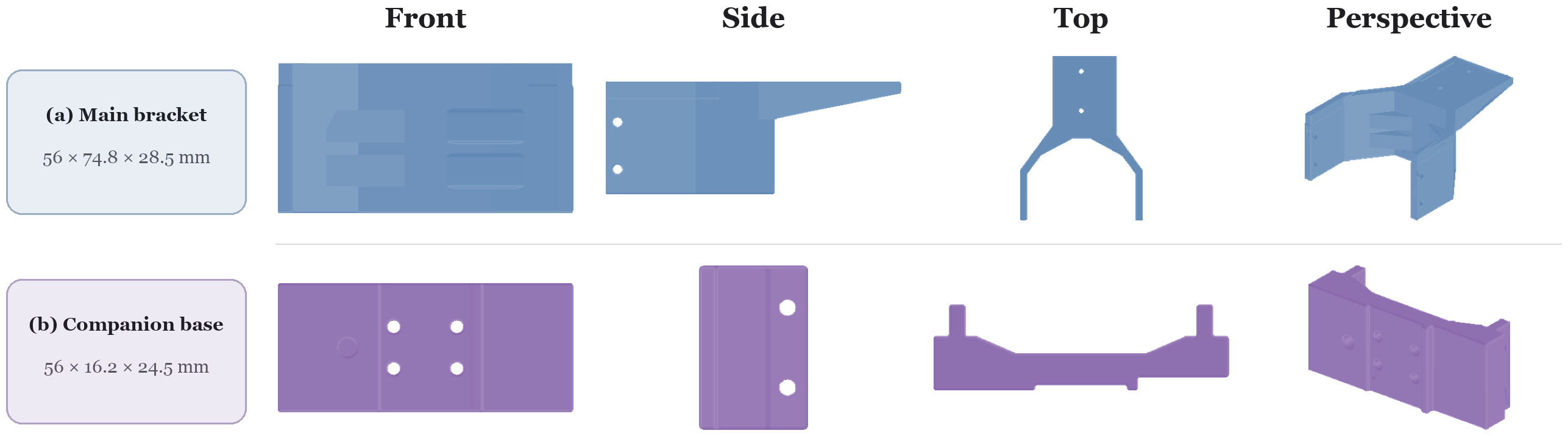}
\caption{Two-part 3D-printed mount for the GelSight Mini. (a) Main bracket and (b) companion base; the columns show front, side, top, and perspective views. Both parts form one gripper-mounted sensor assembly.}
\label{fig:app_gelsight_mount}
\end{figure}

\paragraph{Hardware and Observations.}
Our setup comprises a PiPER six-degree-of-freedom arm, a Pika parallel gripper, and two GelSight Mini sensors mounted on opposing fingers (Figure~\ref{fig:app_hardware}). Following the gripper-mounted optical tactile setup used in AnyTouch~2~\citep{feng2026anytouch2}, each sensor is attached with a compact two-piece 3D-printed assembly (Figure~\ref{fig:app_gelsight_mount}). The policy observes one task-dependent RGB stream, the two synchronized tactile streams, and robot state. Play Mahjong, Pick Up Chips, and Insert USB use a fixed third-person camera, whereas Wipe Board uses a wrist camera to keep the board in view throughout contact. The GelSight cameras stream marker-preserving RGB images at 640$\times$480 and 25 Hz. Before policy inference, RGB observations are resized to 480$\times$270 and tactile observations to 320$\times$240. PiPER provides six arm-joint positions and one gripper coordinate. We map this physical 7-dimensional state to the shared 8-dimensional interface as $[q_1,\ldots,q_6,0,g]$, where the zero is a dummy seventh arm joint matching the UniVTAC convention. The same dummy dimension is removed from predicted actions before sending commands to PiPER.

\paragraph{Demonstrations and Deployment.}
Demonstrations are collected through the Pika interface at a uniform 20 Hz timeline. Camera, tactile, and joint-state messages are associated by nearest timestamps; samples with excessive inter-stream skew are rejected during dataset preparation. The policy predicts a 50-step chunk in the mixed action representation used by FTP-1~\citep{yuan2026ftp}: arm-joint commands are expressed as deltas from the current state, while the gripper command remains absolute. Before execution, the joint deltas are converted back to absolute PiPER joint targets and streamed at 20 Hz. Table~\ref{tab:app_real_execution_protocol} summarizes the task-wise execution protocol. Pick Up Chips executes 25 actions before refreshing its observations, whereas the other tasks execute the full 50-step chunk. Evaluation rollouts receive no human intervention after execution begins. Emergency stops and infrastructure failures are recorded separately from task failures; the affected rollouts are rerun using the same reset protocol.

\begin{table}[htbp]
\centering
\caption{Real-robot execution protocol for \simpletouch{}. ``Predicted'' and ``Executed'' denote the action-prediction and execution horizons, respectively.}
\label{tab:app_real_execution_protocol}
\apptablecaptionspace
\small
\setlength{\tabcolsep}{7pt}
\begin{tabular}{lccccc}
\toprule
Task & Demonstrations & RGB view & Predicted & Executed & Eval. trials\\
\midrule
Play Mahjong & 50 & Fixed & 50 & 50 & 20\\
Wipe Board & 50 & Wrist & 50 & 50 & 20\\
Pick Up Chips & 50 & Fixed & 50 & 25 & 20\\
Insert USB & 50 & Fixed & 50 & 50 & 20\\
\bottomrule
\end{tabular}
\end{table}

\paragraph{Task Protocols.}
Table~\ref{tab:app_real_protocol} summarizes the task-level recipe used for both demonstration collection and evaluation. We vary only conditions that preserve the semantic goal and remain within the demonstrated setup. Success is judged as a binary task outcome at the end of each rollout.

\begin{table}[htbp]
\centering
\caption{Real-robot task protocols. Each policy is trained with 50 demonstrations and evaluated for 20 rollouts per task.}
\label{tab:app_real_protocol}
\apptablecaptionspace
\small
\setlength{\tabcolsep}{4pt}
\begin{tabular}{p{0.13\linewidth}p{0.10\linewidth}p{0.34\linewidth}p{0.31\linewidth}}
\toprule
Task & RGB view & Initial-condition variation & Binary success criterion\\
\midrule
Play Mahjong & Fixed & The grasped tile is either a bird tile or a one-dot tile; its pose within the gripper is varied. A two-second tactile observation is recorded at the start. & The bird tile is discarded to the table center, or the one-dot tile is placed in front of the robot, followed by a stable release.\\
Wipe Board & Wrist & Board pose and angle, writing size, and eraser placement in the gripper are varied. & The marked region is wiped clean while the eraser remains securely grasped.\\
Pick Up Chips & Fixed & The chip and supporting bottle-cap positions are varied. & The chip is grasped and lifted clear without being dropped or visibly broken.\\
Insert USB & Fixed & USB placement in the gripper is varied while its insertion orientation is kept consistent with the port. & The connector is fully seated in the port and remains inserted after release.\\
\bottomrule
\end{tabular}
\end{table}

\section{Baselines and Controlled Variants}
\label{app:variants}
\subsection{Baseline Methods}
We compare with seven policies spanning vision-only imitation learning, pretrained VLA adaptation, visuotactile representation learning, and tactile policy pretraining.

\paragraph{ACT.}
Action Chunking with Transformers (ACT)~\citep{zhao2023act} is a task-trained visuomotor imitation policy that predicts temporally coherent action chunks from visual observations and proprioceptive state. It receives no tactile input and provides a vision-only reference without policy pretraining.

\paragraph{$\pi_{0.5}$.}
$\pi_{0.5}$~\citep{intelligence2025pi05} is a pretrained VLA policy that couples a PaliGemma visual-language backbone with a flow-matching action expert. In our comparison it is adapted to each task using the available demonstrations but receives no tactile observation, isolating the benefit of adding touch to a strong pretrained VLA.

\paragraph{VITaL.}
VITaL~\citep{george2024vital} performs a separate visuotactile pretraining stage before downstream imitation learning. By learning from paired visual and tactile observations, it supplies the policy with an aligned cross-modal representation and represents the explicit-alignment route to tactile policy learning.

\paragraph{UniVTAC-ACT.}
UniVTAC-ACT~\citep{chen2026univtac} is the tactile ACT baseline introduced with the UniVTAC benchmark. It augments the ACT policy with the tactile-centric representation learned by the UniVTAC Encoder, allowing action prediction to condition on both visual and tactile observations.

\paragraph{Tactile-VLA.}
Tactile-VLA~\citep{huang2025tactilevla} connects tactile observations to the physical-interaction knowledge of a pretrained VLA. It combines multimodal fusion with tactile-aware reasoning and hybrid position-force control, providing a task-level route for adapting a pretrained VLA to contact-rich manipulation.

\paragraph{FTP-$\pi_{0.5}$.}
FTP-$\pi_{0.5}$~\citep{yuan2026ftp} uses the FTP-1 architecture initialized from $\pi_{0.5}$ but omits tactile policy pretraining. Its tactile pathway is learned only from the task demonstrations, making it the closest evaluated baseline for testing whether an existing tactile-policy architecture is sufficient without its large-scale pretraining stage.

\paragraph{FTP-1.}
FTP-1~\citep{yuan2026ftp} is a generalist tactile policy pretrained on heterogeneous tactile manipulation data spanning multiple sensors and embodiments. It is subsequently adapted to each downstream task and serves as our principal comparison to large-scale tactile policy pretraining.

\subsection{Representation and Prediction Ablations}
The controlled variants in Figure~\ref{fig:ablations} and Table~\ref{tab:predictive_ablation} probe different parts of the recipe. Their definitions are separated in Table~\ref{tab:app_variants} so that removing tactile inputs, changing the encoder, and removing prediction supervision are not conflated.

\begin{table}[htbp]
\centering
\caption{Definitions of the reported representation and prediction ablations.}
\label{tab:app_variants}
\apptablecaptionspace
\small
\begin{tabular}{>{\raggedright\arraybackslash}p{0.18\linewidth}>{\raggedright\arraybackslash}p{0.31\linewidth}>{\raggedright\arraybackslash}p{0.42\linewidth}}
\toprule
Variant & Controlled change & Exact setting\\
\midrule
Vision-only & Remove touch from the policy & Disable the tactile encoder and tactile expert while retaining the task data and $\pi_{0.5}$ action backbone.\\
Token count & Vary current-touch spatial resolution & Use the T3 CLS token alone (1), or CLS plus adaptively pooled $3{\times}3$ (10), $6{\times}6$ (37), or $8{\times}8$ (65) spatial tokens; 197 retains CLS plus the complete $14{\times}14$ grid. Counts are per sensor.\\
VAE & Replace the tactile-specific representation & Use a frozen Wan2.2 video VAE from Fast-WAM. The two RGB tactile images are concatenated, encoded as one latent frame, and rearranged into 98 tokens of width 192 before projection to the tactile expert. No CLS token or spatial pooling is used.\\
Trainable T3 & Update the tactile encoder & Optimize T3 jointly with the policy from current tactile images; keep future-prediction targets fixed to cached features from the original pretrained T3.\\
w/o Tac. Pred. & Remove predictive supervision & Set the future-tactile loss weight to zero while retaining the full current-touch pathway and tactile expert.\\
Single-horizon & Use one future prediction target & Predict CLS plus the same pooled $6{\times}6$ spatial grid (37 tokens per sensor) only at $t{+}50$, rather than at ten offsets $t{+}5,\ldots,t{+}50$.\\
\simpletouch{} & Full proposed recipe & Use all 197 current T3 tokens per sensor and multi-horizon future prediction with a $6{\times}6$ pooled spatial grid.\\
\bottomrule
\end{tabular}
\end{table}

All token-count variants preserve the multi-horizon targets, loss weight, and optimization recipe, so this sweep changes only the spatial granularity of the current tactile context. Adaptive average pooling preserves the global CLS token while reducing the $14{\times}14$ patch grid. The vision-only ablation uses the same $\pi_{0.5}$ model and training recipe as the corresponding baseline in Table~\ref{tab:main}. The only difference is the execution horizon for the insertion tasks: the main-table baseline uses 50 actions, whereas the ablation executes 25 actions before requesting a new observation.

The encoder comparison keeps the expert architecture and predictive objective unchanged. The VAE variant uses the pretrained Wan2.2 VAE checkpoint and its deterministic posterior mean, with the codec kept fixed during policy training. The trainable-T3 variant instead updates only the current-touch encoder path against immutable pretrained-T3 future targets, preventing the auxiliary target itself from drifting. The single- and multi-horizon variants retain frozen T3 features and use the same CLS-plus-$6{\times}6$ pooled target at each horizon. Single-horizon supervises only $t{+}50$, whereas the full method supervises ten offsets from $t{+}5$ to $t{+}50$, varying temporal supervision while holding per-horizon spatial resolution fixed.

\subsection{Demonstration-Budget Protocol}
Figure~\ref{fig:data_efficiency} compares 10, 20, 50, and 100 task demonstrations for Put Bottle and Insert Hole. For a budget of $N$, training uses the deterministic prefix of episodes $0,\ldots,N{-}1$; the subsets are therefore strictly nested. State and action normalization statistics are recomputed from the selected subset, whereas the pretrained $\pi_{0.5}$ and T3 weights and the fixed tactile preprocessing statistics are shared across budgets. The budget counts task demonstrations only and does not include upstream VLA or tactile-representation pretraining data.

We scale optimization duration with the task data: the 10-, 20-, 50-, and 100-demonstration settings train for 4k, 8k, 20k, and 40k updates, respectively, with a global batch size of 64. Warmup remains 5\% of training, followed by the same cosine schedule. Model architecture, future-tactile loss, camera protocol, and tail handling remain unchanged. Each point is evaluated with the same task-specific inference protocol as the main results.

\subsection{Joint, IDM, and Fast Configurations}
The three formulations preserve the same backbone, tactile encoder, tactile expert, multi-horizon targets, loss weights, task data, and repeat-tail supervision. They differ only in whether action generation is conditioned on future tactile variables. Their exact information flow is summarized in Table~\ref{tab:app_wam_modes}; the measured accuracy and efficiency are reported in Appendix~\ref{app:efficiency_protocol}.

\begin{table}[htbp]
\centering
\caption{Training and inference protocols for the three tactile-prediction formulations.}
\label{tab:app_wam_modes}
\apptablecaptionspace
\small
\begin{tabular}{>{\raggedright\arraybackslash}p{0.12\linewidth}>{\raggedright\arraybackslash}p{0.42\linewidth}>{\raggedright\arraybackslash}p{0.37\linewidth}}
\toprule
Mode & Training-time action conditioning & Inference\\
\midrule
Joint & Action tokens read the K/V cache produced by the same noisy future-tactile branch. Action and tactile flow times are sampled independently, and future-tactile queries never read action tokens. & Future touch and actions are updated together for 10 Euler steps.\\
IDM & Actions read a separate teacher-forced future-tactile branch. Each sample uses the clean target or, with probability 0.5, an independently noised target; this branch is separate from the auxiliary future-loss branch. & Future touch is denoised for 10 steps, cached at its clean endpoint, and then conditions 10 action-denoising steps.\\
Fast & The action branch reads only visual-language, state, and current-tactile context. Future tactile prediction is an auxiliary training objective and cannot condition action tokens. & The cached current context directly conditions 10 action-denoising steps; no future tactile latent is generated.\\
\bottomrule
\end{tabular}
\end{table}

For all three modes, current tactile features are prefetched once into a layerwise cache. Future-tactile flow matching uses the same Beta$(1.5,1)$ time distribution as action flow matching, and both branches use the same multi-horizon T3 targets. This controlled construction attributes the differences in Figure~\ref{fig:efficiency} to the prediction formulation rather than to representation capacity or supervision.

\section{Additional Analyses and Efficiency}
\label{app:figures}

\subsection{Additional Future-Tactile Visualizations}
\label{app:additional_tactile_visualizations}
\begin{figure}[t]
    \centering
    \includegraphics[width=\linewidth]{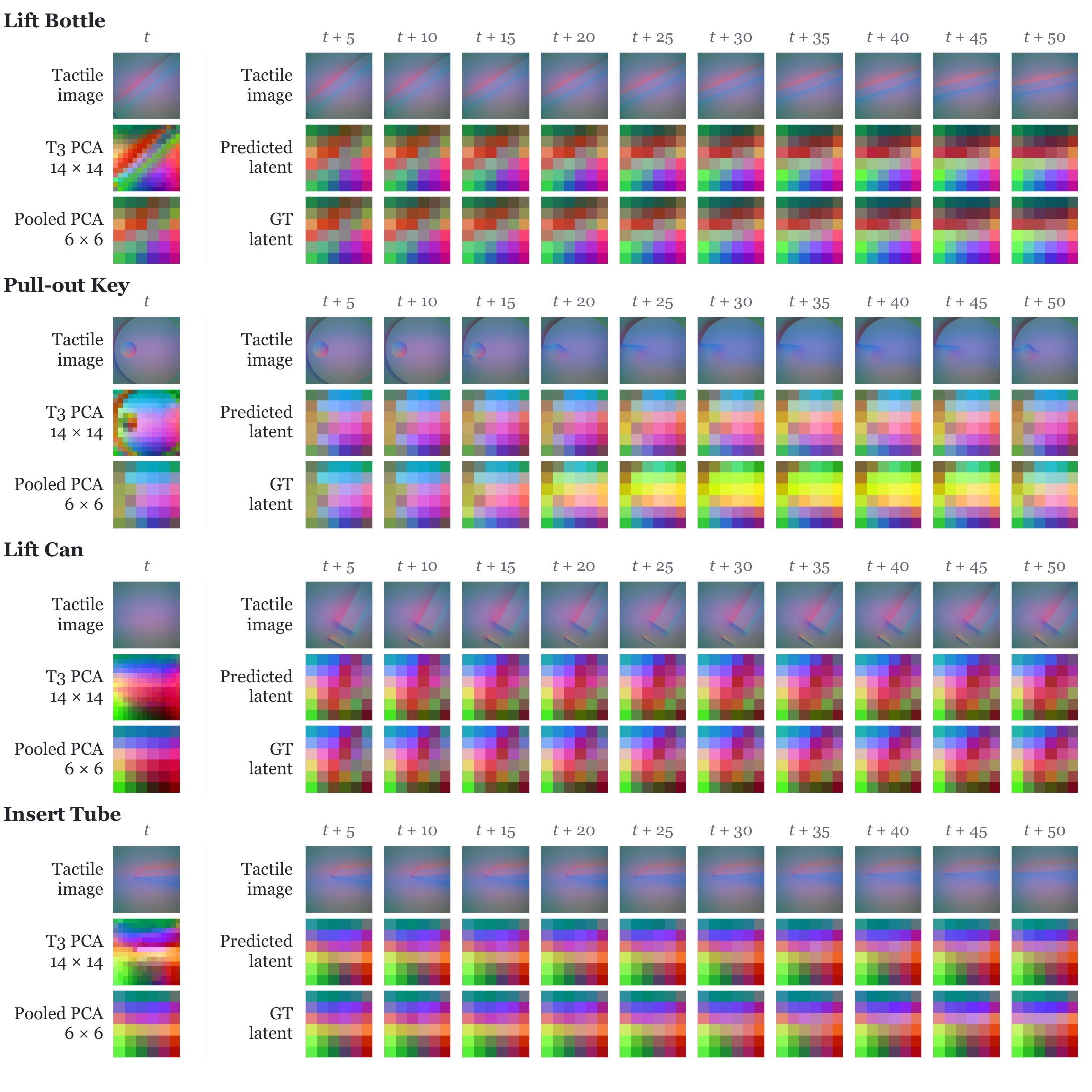}
    \caption{Additional multi-horizon future-tactile predictions for Lift Bottle, Pull-out Key, Lift Can, and Insert Tube. For each task, the current tactile image and its T3 representations appear on the left; subsequent columns compare the observed tactile image, predicted pooled latent, and ground-truth pooled latent at 5-step offsets through $t+50$.}
    \label{fig:app_tactile_remaining}
\end{figure}

Figure~\ref{fig:tactile_prediction} compares observed tactile images with PCA visualizations of predicted and ground-truth future tactile latents on Put Bottle and Insert Hole. Figure~\ref{fig:app_tactile_remaining} extends the same analysis to Lift Bottle, Pull-out Key, Lift Can, and Insert Tube, jointly covering contact formation, sustained contact, object motion, and insertion correction. Within each sequence, predicted and ground-truth latents use the same PCA basis fitted to the ground-truth latents, and the displayed color limits are fixed from the ground-truth sequence. Across these task families, the predictions qualitatively follow the principal spatial evolution of contact over multiple horizons rather than an arbitrary auxiliary target.

\subsection{Execution Horizon and Feedback Frequency}
\label{app:execution_horizon}
The policy is trained to predict a 50-action chunk in every task. At deployment, the execution horizon determines how many actions are applied before the policy observes the scene again and replans. Based on task dynamics, we specify H50 for the four non-insertion tasks and H25 for the two insertion tasks before evaluation, without validation- or test-set tuning. Table~\ref{tab:execution_horizon} then evaluates the same task policies under both horizons as a sensitivity analysis; it isolates feedback frequency at inference rather than changing the training target or checkpoint.

\begin{table}[htbp]
\centering
\caption{Effect of execution horizon. Every policy is trained with a 50-action prediction horizon; H25 and H50 execute 25 and 50 actions, respectively, before the next observation. Success rates are reported in percent.}
\label{tab:execution_horizon}
\apptablecaptionspace
\small
\setlength{\tabcolsep}{2pt}
\renewcommand{\arraystretch}{1.08}
\begin{tabular}{@{}*{8}{>{\centering\arraybackslash}p{0.116\linewidth}}@{}}
\toprule
Execution & Lift Bottle & Pull-out Key & Lift Can & Put Bottle & Insert Hole & Insert Tube & Avg.\\
\midrule
H50 & \textbf{100} & \textbf{62} & \textbf{68} & \textbf{50} & 58 & 88 & \textbf{71.0}\\
H25 & \textbf{100} & 52 & 42 & 39 & \textbf{85} & \textbf{100} & 69.7\\
\bottomrule
\end{tabular}
\end{table}

The preferred execution horizon depends on how quickly useful feedback changes. Lift, extraction, and placement generally benefit from H50: their successful motion is comparatively coherent over the predicted chunk, and executing a longer segment avoids interrupting that motion with unnecessary resampling. Lift Bottle is insensitive in this comparison, whereas the other three non-insertion tasks lose 10--26 points with H25. Insertion reverses the trend. Small changes in contact geometry can alter the required correction direction, so observing touch twice as frequently helps the policy respond before a misalignment becomes a jam, overshoot, or slip. H25 improves Insert Hole and Insert Tube by 27 and 12 points, respectively. These results support the pre-specified task-level rule used in Table~\ref{tab:main} without modifying the trained policies.

\subsection{Training and Inference Efficiency}
\label{app:efficiency_protocol}
Table~\ref{tab:app_efficiency} provides the task-wise measurements underlying Figure~\ref{fig:efficiency}. Training time is the mean wall-clock time per optimization step with eight H20 GPUs and a global batch size of 64, measured after initialization and data-cache preparation and excluding checkpoint saving. Inference latency is batch-one wall-clock time from synchronized model inputs to one complete action chunk, after warm-up and with GPU synchronization. It includes tactile encoding, context construction, and all model forward passes required by each formulation, and excludes simulator or robot execution and video recording. All methods use ten Euler integration steps, and inference is measured on one A100 GPU.

\begin{table}[htbp]
\centering
\caption{Task-wise training and inference efficiency for Joint, IDM, and Fast.}
\label{tab:app_efficiency}
\apptablecaptionspace
\small
\setlength{\tabcolsep}{4pt}
\begin{tabular}{@{}ll*{4}{>{\centering\arraybackslash}p{0.12\linewidth}}@{}}
\toprule
Task & Formulation & \shortstack{Training\\(s/step)} & \shortstack{Latency\\(s/query)} & \shortstack{SR\\(\%)} & \shortstack{SR/latency\\(\%/s)}\\
\midrule
Pull-out Key & Joint & 1.891 & 0.946 & 56 & 59.2\\
 & IDM & 2.275 & 1.642 & 51 & 31.1\\
 & Fast & 1.862 & 0.614 & 62 & 101.0\\
Put Bottle & Joint & 1.890 & 1.058 & 46 & 43.5\\
 & IDM & 2.279 & 1.444 & 38 & 26.3\\
 & Fast & 1.856 & 0.632 & 50 & 79.1\\
Insert Hole & Joint & 1.898 & 1.049 & 80 & 76.3\\
 & IDM & 2.280 & 1.372 & 82 & 59.8\\
 & Fast & 1.868 & 0.603 & 85 & 141.0\\
Insert Tube & Joint & 1.897 & 1.026 & 99 & 96.5\\
 & IDM & 2.285 & 1.394 & 98 & 70.3\\
 & Fast & 1.866 & 0.643 & 100 & 155.5\\
\midrule
Mean & Joint & 1.894 & 1.020 & 70.3 & 68.9\\
 & IDM & 2.280 & 1.463 & 67.3 & 46.9\\
 & Fast & 1.863 & 0.623 & 74.3 & 119.1\\
\bottomrule
\end{tabular}
\end{table}

Fast has the lowest mean training time and inference latency while obtaining the highest mean success. Relative to Joint and IDM, it reduces inference latency by 38.9\% and 57.4\%, respectively; its mean SR/latency is 1.73 and 2.54 times as high, respectively. The task-level results show that this efficiency does not come from sacrificing contact-sensitive control.

\section{Limitations and Future Directions}
Our current real-robot evaluation covers parallel-jaw grippers equipped with vision-based tactile sensors; dexterous hands and other tactile sensor types remain unevaluated. Future work should assess whether the approach remains effective across these broader hardware configurations. The present study focuses on tactile understanding, leaving high-frequency tactile feedback and force-based control for future exploration.

\end{document}